\documentclass[journal]{IEEEtran}
\usepackage{amsmath,graphicx,amsthm,amssymb,bbm, subfigure, algorithm, bm, algorithmic, cite,xcolor,fontenc,subcaption,booktabs}
\usepackage{adjustbox}
\usepackage{hyperref}
\usepackage{textcomp}
\usepackage[normalem]{ulem}
\usepackage{cases}
\usepackage{hyperref,url}
\newtheorem{Theorem}{Theorem}
\newtheorem{Lemma}{Lemma}
\newtheorem{Assumption}{Assumption}
\theoremstyle{definition}
\newtheorem{remark}[Theorem]{Remark}
\newcommand\mc{\mathcal}
\newcommand\bd{\boldsymbol}

\begin{document}
\title{Diffusion Stochastic Optimization for Min-Max Problems}
%
%
%

\author{Haoyuan Cai{$^*$}, \IEEEmembership{Student Member, IEEE}, Sulaiman A. Alghunaim{$^\bullet$}, \IEEEmembership{Member, IEEE}, and Ali H. Sayed{$^*$}, \IEEEmembership{Fellow, IEEE}\\
$^\bullet$Kuwait University, Kuwait\\
$^*$\'Ecole Polytechnique F\'ed\'erale de Lausanne, Switzerland
\thanks{
A short version {\color{black} of this work without derivations will appear in the conference publication \cite{hao2024}.}
Emails: haoyuan.cai@epfl.ch, sulaiman.alghunaim@ku.edu.kw, ali.sayed@epfl.ch}}


\maketitle

\begin{abstract}
\textcolor{black}{
The optimistic gradient method is useful in addressing} minimax optimization \textcolor{black}{problems}.
Motivated by 
the observation that the conventional stochastic \textcolor{black}{version} suffers from \textcolor{black}{the need for a}
 large 
batch size  \textcolor{black}{on the order of} $\mathcal{O}(\varepsilon^{-2})$ \textcolor{black}{to achieve} an $\varepsilon$-stationary solution, we introduce and analyze 
a new formulation termed Diffusion Stochastic Same-Sample Optimistic Gradient (DSS-OG). We prove its convergence and 
\textcolor{black}{resolve} the large batch issue
by establishing
a tighter upper bound, under  
the more general \textcolor{black}{setting of} nonconvex Polyak-Lojasiewicz (PL) \textcolor{black}{risk functions}.
We also extend the applicability of the proposed method to  {\color{black}the} distributed \textcolor{black}{scenario}, where agents communicate with their neighbors via a left-stochastic protocol.
To implement DSS-OG,
we can query the stochastic gradient oracles
in parallel with some extra memory overhead, resulting in a complexity comparable to its conventional counterpart. To demonstrate the efficacy of \textcolor{black}{the} proposed algorithm, we conduct tests by training generative adversarial networks.
\end{abstract}


%
\IEEEpeerreviewmaketitle

\section{Introduction}
\label{sec:introduction}
\IEEEPARstart{M}{inimax}  optimization takes center stage in many machine learning applications,  including generative adversarial networks
(GANs) \cite{goodfellow2014generative}, adversarial machine learning \cite{madry2017towards},  and reinforcement
learning \cite{dai2018sbeed}.

Minimax problems {\color{black} take the form:
\vspace{-0.5em}
\begin{align} \label{local_risk_centralized}
    \underset{x \in \mathbb{R}^{M_1}}{\min} 
    \underset{y \in \mathbb{R}^{M_2}}{\max}  \ J(x, y) 
\end{align} 
where $x$ and $y$ are the primal and dual variables, respectively, and $J(x,y)$ is the risk (objective) function. This problem is} typically approached from various perspectives.
In the optimization \textcolor{black}{literature},
assuming certain structure 
\textcolor{black}{for} the risk function over the variables
is \textcolor{black}{prevalent}.
For instance, 
the risk function can be assumed
(strongly) convex in the primal variable and (strongly) concave in the dual variable
\cite{NEURIPS2022_5e2ed801, zhang2022lower} or it can be assumed one-sided nonconvex/Polyak-Lojasiewicz (PL) \cite{NEURIPS2022_Tight, yang2022faster,lin2020gradient}
in \textcolor{black}{the} primal variable and one-sided strongly concave/PL in \textcolor{black}{the}
 dual variable.
Some works
assume a two-sided PL condition\cite{yang2020global}.
It is noted that the gradient operator of a (strongly) convex or (strongly) concave objective 
is also 
(strongly) monotone; therefore, 
the optimization problem \textcolor{black}{is} generalized \textcolor{black}{to} a
variational inequality (VI) problem in some settings. Solving
VI \textcolor{black}{problems} heavily relies on the concept of 
\textcolor{black}{``operators"} and \textcolor{black}{their} associated properties.
The goal of \textcolor{black}{a} VI problem is to determine the solution at which the gradient operator satisfies  variational stability conditions, such as the Stampacchia variational inequality (SVI), which \textcolor{black}{is} a first-order necessary optimality condition
\cite{kinderlehrer2000introduction,komlosi1999stampacchia}.
{In a similar vein}, the Minty variational inequality (MVI) condition and its weak version, 
which 
include a subclass of nonmonotone problems,
has also gained popularity in recent years 
\cite{dou2021one,beznosikov2022decentralized,bohm2023solving,diakonikolas2021efficient,pethick2023solving}.
{The} settings mentioned above represent the tractable conditions {for} which convergent algorithms are known to exist.
On the other hand,
the nonconvex-nonconcave formulation,
{which is at the core} 
of a wide class of minimax objectives, 
{remains} intractable \cite{daskalakis2021complexity}.
This fact corroborates the significant difficulty encountered in training  GANs, as they often involve nonconvex-nonconcave objectives.


The optimistic gradient (OG) method and its variants have attracted significant interest for their proven convergence in  wide formulations of \eqref{local_risk_centralized}
\cite{korpelevich1976extragradient,bohm2023solving,beznosikov2022decentralized, NEURIPS2022_Tight, dou2021one, liu2020decentralized}.
However, {a good part of the} theory from this body of literature  {breaks down} when the stochastic setting 
is taken into consideration {\color{black} where only a noisy random estimate of the gradient is available.}
For instance, 
the work  \cite{NEURIPS2022_Tight}
showed that
employing
 a batch size \textcolor{black}{on the order of}  $\mathcal{O}(\varepsilon^{-2})$ 
 is necessary
{to estimate the gradient and in order to control} the gradient noise for stochastic OG method in a nonconvex-strongly concave  (SC) setting. {Here, the scalar} $\varepsilon$  denotes 
{the desired level of accuracy in the solution, which is generally a very small number. Similarly,} the work  \cite{liu2020decentralized} 
established \textcolor{black}{similar} results with a batch size  $\mathcal{O}(\varepsilon^{-4})$
to achieve convergence in the stochastic MVI \textcolor{black}{setting}.
These are not isolated cases;
a similar need for using large batch sizes appears in the works \cite{ beznosikov2022decentralized, dou2021one, lin2020gradient, bohm2023solving}. 
The reliance on batch sizes \textcolor{black}{that are} inversely proportional to 
$\varepsilon$ is a \textcolor{black}{serious} theoretical limitation since it
 rules out the freedom of \textcolor{black}{using} small \textcolor{black}{batches}. Meanwhile, 
empirical results showed that stochastic OG may not be subject to such a restrictive requirement
\cite{daskalakis2018training,liu2020decentralized}. 

In this work, we investigate distributed {\color{black} stochastic} minimax problems where a network of $K$ agents cooperate to solve the following problem:
\vspace{-1.5em}
\begin{subequations}
\begin{align} 
    \underset{x \in \mathbb{R}^{M_1}}{\min} 
    \underset{y \in \mathbb{R}^{M_2}}{\max}  \ J(x, y) &= \sum_{k = 1}^{K}p_k
    J_k(x, y)   \label{risk}
\\
\text{where}  \quad    J_k(x, y)  &\triangleq
\mathbb{E}_{\boldsymbol{\xi}_k} Q_k(x,y; \boldsymbol{\xi}_k) \label{local_risk}
\end{align} 
\end{subequations}
Here, $x, y$ are the global primal and dual variables, the function
$Q_k(x,y; \boldsymbol{\xi}_k)$ is the local 
{\em stochastic} loss \textcolor{black}{evaluated at} the sample  $\boldsymbol{\xi}_k$. 
The scalars
$p_k$ are positive weights  
satisfying $\sum_{k=1}^{K} p_k = 1$. We assume that each agent $k$ only has access to its local data set $\{\boldsymbol{\xi}_k\}$. {\color{black} We consider a fully distributed (decentralized) setting where the agents are 
}
\textcolor{black}{connected by a graph topology and each agent} can only share information with \textcolor{black}{its immediate} neighbors \cite{sayed2014adaptation}.
We approach  \eqref{risk}-\eqref{local_risk} from the perspective of optimization, adopting assumptions that are more general than the nonconvex-SC formulation. {\color{black} Instead of assuming the dual risk function to be SC in the dual variable, we consider the weaker PL-condition \cite{karimi2016linear}.} Specifically, we adopt the following assumption.
\begin{Assumption}[\textbf{Risk function}] \label{muPL}
We assume each local {\color{black} risk} function
$J_k(x, y)$
is  nonconvex in $x$ while 
$-J(x, y)$ is $\nu$-PL in $y$, 
i.e., for any $x$ and $y$ in the domain of $J(x,y)$, it holds that
\begin{equation} 
 \|\nabla_{y} J(x, y)\|^2
 \ge 2\nu(\underset{y} {\max} \ J(x, y) -  J(x, y))
\end{equation}
where $\nu$ is some strictly positive constant. 
$\hfill{\square}$
\end{Assumption}

Under this assumption,
we are interested in answering the following questions:
\textbf{Q1:} {\em Can the reliance on large batch sizes be eliminated in the stochastic OG method?}
\textbf{Q2:} {\em What are the performance guarantees of the
\textcolor{black}{resulting}
batch-flexible stochastic OG over \textcolor{black}{networks}?}
\subsection{Related works}
Minimax optimization problems \textcolor{black}{have been largely studied} under the {\em single} node setting \cite{zhang2022lower,zhang2022near, NEURIPS2022_5e2ed801, lin2020gradient, chavdarova2019reducing, liu2022quasi, golowich2020last,gorbunov2022last,ryu2019ode,yoon2021accelerated}.
The fundamental setting assumed within these works
is the strong convexity-strong concavity/strong monotonicity.
For this setting,
the work \cite{zhang2022lower} established the tight complexity bound of the first-order algorithm in finding a near-optimal solution.
The work \cite{NEURIPS2022_5e2ed801}
devised the first optimal algorithm that matches this bound.
The work \cite{zhang2022near}
showed that the alternating gradient 
\textcolor{black}{descent-ascent} (AGDA) method  
converges faster than
its simultaneous counterpart.
The work \cite{chavdarova2019reducing}  proposed a stochastic variance-reduced extragradient (EG) method, which showed improved performance in training  GANs.
The quasi-Newton methods \textcolor{black}{have also been} proposed to achieve a faster convergence rate in finding the saddle point \cite{liu2022quasi}.
Dropping  
``strongly" setting from both sides of strong
convexity-strong concavity
brings about convergence issues \textcolor{black}{where} simultaneous GDA \textcolor{black}{will} suffer from divergent rotation while AGDA \textcolor{black}{will get} stuck in a limit cycle rather than converge to a \textcolor{black}{limit} point \cite{zhang2022near, gidel2019negative}.
For the convex-concave setting,
the work \cite{ryu2019ode}  established the sublinear rate of
$\mc{O}(\frac{1}{T})$  
for the {\em best}-iterate of the stochastic OG method, namely the best iterate produced over running $T$ iterations.
Achieving last-iterate convergence
is more difficult;
this aspect of OG was not addressed until the work \cite{gorbunov2022last}.
Furthermore,
the  work
\cite{golowich2020last}
showed that the
{\em average}-iterate of EG, namely the average of all iterates produced over iterations,  has a  convergence rate of
$\mc{O}(\frac{1}{T^2})$ which is
faster than its last-iterate.
The 
work \cite{yoon2021accelerated} also \textcolor{black}{showed} anchoring EG has a convergence rate
of $\mc{O}(\frac{1}{T^2})$
 without considering stochasticity.
There is another line of research that investigates the structured nonmonotone problem, particularly under the premise of the (weak) Minty variational inequality--- see \cite{diakonikolas2021efficient,  pethick2023solving, bohm2023solving}.

In recent years, there has been an increasing interest
towards the
stochastic nonconvex (one-sided) minimax optimization \textcolor{black}{problem}.
The work 
\cite{lin2020gradient} 
\textcolor{black}{established} the convergence of
stochastic GDA (SGDA) 
in finding the 
primal solution
under a nonconvex-SC
formulation,
with the use of batch size  $\mathcal{O}(\varepsilon^{-2})$ {\color{black} to reach $\varepsilon$ error accuracy.}
The work  \cite{luo2020stochastic}
proposed to incorporate a concave maximizer 
and variance reduction \textcolor{black}{into} SGDA, which is computationally costly
since it performs multi-loop \textcolor{black}{steps} in updating the dual variable.
The work \cite{yang2020global} studied the two-sided PL setting and \textcolor{black}{established} the linear convergence of variance-reduced stochastic AGDA (SAGDA).
Additionally, the work \cite{yang2022faster} showed that employing two iid samples for SAGDA obtains $\mathcal{O}(\frac{1}{T^{1/2}})$ convergence rate
for the primal solution in 
nonconvex-PL setting. 
However, vanilla AGDA 
does not converge in bilinear scenarios\cite{gidel2019negative}, which casts doubt on its ability to solve more complex settings.
Exploiting 
the convergence of the backbone minimax algorithms across various settings, say OG/EG, 
is always of independent interest. Unfortunately,
the recent work 
\cite{NEURIPS2022_Tight}
\textcolor{black}{showed} a less than ideal result under the nonconvex-SC formulation as the large batch issue \textcolor{black}{persists}. This fact strongly motivates us to pursue better
 results that theoretically address this limitation.

Over the past decade,
distributed optimization has become an increasingly important field due to its flexibility in processing large volume and physically dispersed data \cite{sayed2014adaptation,sayed2022inference}.
The distributed minimax optimization 
\textcolor{black}{scenario} remains an open field to be exploited \textcolor{black}{since} there are \textcolor{black}{many} diverse settings.
\textcolor{black}{Some} distributed nonconvex minimax works \textcolor{black}{appear in} \cite{xian2021faster,gao2022decentralized, chen2022simple, huang2023near, beznosikov2022decentralized, liu2020decentralized}.
 The work \cite{xian2021faster} \textcolor{black}{showed} the convergence rate of  $\mathcal{O}(\frac{1}{T^{2/3}})$ for the primal solution
using STORM momentum-based \cite{cutkosky2019momentum} SGDA, under a nonconvex-SC setting.
The  work \cite{huang2023near} extended the analysis to the nonconvex-PL setting. 
Other works \cite{gao2022decentralized, chen2022simple} introduced variance-reduced SGDA for the nonconvex-SC setting. However,
variance reduction requires an
excessively large batch size
to estimate the gradient 
at the checkpoint, making it hard to tune.
On the other hand,  \textcolor{black}{the} works
\cite{xian2021faster, cutkosky2019momentum,huang2023near,chen2022simple, gao2022decentralized} rely on the Lipschitz continuity of \textcolor{black}{the} stochastic loss gradient to carry out \textcolor{black}{the} analysis.
Relaxing this condition leads to
\textcolor{black}{a} limited applicability of the results \cite{arjevani2023lower}.
\textcolor{black}{The} work \cite{beznosikov2022decentralized} proposed a distributed version of stochastic EG for solving stochastic MVI in a time-varying network. 
The work \cite{liu2020decentralized}
proposed a single-call EG
\cite{popov1980modification} for distributed parallel training of the GANs.
However, both results face \textcolor{black}{the same} large batch size issue.

\vspace{-1em}
\begin{table*}[htbp]  
    \centering
    \begin{tabular}{ccccc} 
    \toprule
    \toprule
     \textbf{Works}    & \textbf{Assumptions} &
     \textbf{Primal  Rate}
     & \textbf{Dual  Rate}& \textbf{Batch size}\\
     \hline 
DPOSG \cite{
 liu2020decentralized}
 & Stochastic MVI, bounded gradient, D-S & 
 $\mc{O}(\frac{1}{T^{1/4}} + \frac{\sigma^2}{\sqrt{KB}})$  &   $\mc{O}(\frac{1}{T^{1/4}} + \frac{\sigma^2}{\sqrt{KB}})$ & 
 $\mc{O}(\varepsilon^{-4})$
     \\ 
 Stochastic EG \cite{beznosikov2022decentralized}    & Stochastic MVI, Lipschitz risk gradient, D-S&
 $\mc{O}(\frac{1}{T^{1/2}} + \frac{\sigma^2}{B})$  &  $\mc{O}(\frac{1}{T^{1/2}} + \frac{\sigma^2}{B})$ & 
 $\mc{O}(\varepsilon^{-2})$  
        \\
 DM-HSGD \cite{xian2021faster}   & Stochastic nonconvex-SC,  Lipschitz loss gradient, D-S&
 $\mc{O}(\frac{1}{T^{2/3}})$  &  -------- & 
 $\mc{O}(\frac{1}{\min\{1, K\epsilon\}})$ 
      \\  
DM-GDA \cite{huang2023near}   & Stochastic nonconvex-PL,  Lipschitz loss gradient, D-S&
 $\mc{O}(\frac{1}{T^{1/3}})$  &  -------- & 
 $\mc{O}(1)$
      \\ 
  DSGDA \cite{gao2022decentralized}   & Finite-sum, nonconvex-SC,  Lipschitz loss gradient, D-S&
 $\mc{O}(\frac{1}{T})$  &  -------- & 
 $\mc{O}(\sqrt{n})$
      \\
\textbf{DSS-OG} (\textbf{this work})   &   Stochastic nonconvex-PL, Lipschitz \textcolor{black}{risk} gradient, L-S & $\mc{O}(\frac{1}{T^{1/2}})$ & $\mc{O}(\frac{1}{T})$ & $\mc{O}(1)$   \\
    \bottomrule
    \bottomrule
    \end{tabular}
    \vspace{-0.5em}
\caption{\small Comparison of the convergence rate of different distributed minimax algorithms. 
The works \cite{xian2021faster ,gao2022decentralized}
seek to find  a approximate solution  $\boldsymbol{x}^\star$ of the primal objective $P(x)$ to defined in \eqref{primalobjective} in the sense that
$\mathbb{E}\|\nabla P(\boldsymbol{x}^\star)\|^2 \le \varepsilon^2$ while
the work \cite{huang2023near} uses the criterion $\mathbb{E}\|\nabla P(\boldsymbol{x}^\star)\| \le \varepsilon$.
This work and \cite{ liu2020decentralized, beznosikov2022decentralized} seek to find the approximate solution $(\boldsymbol{x}^\star, \boldsymbol{y}^\star)$  under the convergence criterion 
$\mathbb{E}\| \nabla_{x} J({\boldsymbol{x}}^\star, {\boldsymbol{y}}^\star)\|^2
\le \varepsilon^2, \mathbb{E}\| \nabla_{y} J({\boldsymbol{x}}^\star, {\boldsymbol{y}}^\star)\|^2
\le \varepsilon^2$.
The above table shows the convergence rate of the primal and dual variables after running $T$ iterations, respectively.
    D-S: doubly stochastic,
    L-S: left stochastic,
    $T$: number of iterations,  $B$: batch size employed to achieve an $\varepsilon$-stationary point, $\sigma^2$: gradient noise variance, $n$: number of total samples in finite-sum setting. 
    }
    \label{comparison}
\end{table*}
\subsection{Main Contributions} Our main contributions \textcolor{black}{in} this work
are as follows: 
\begin{itemize}
    \item We introduce and analyze a new variant of 
    stochastic OG that achieves a convergence rate of 
$\mathcal{O}(\frac{1}{T^{1/2}})$ in finding the primal solution under the nonconvex-PL \textcolor{black}{setting}. Our analysis improves the existing performance bound by addressing the issue of large batch size \cite{NEURIPS2022_Tight}.
We also provide insights into theoretical limitations that prevent the conventional stochastic OG from achieving better results.

\item 
\textcolor{black}{We} extend the applicability of stochastic OG to a fully distributed setting.
We consider left-stochastic combination matrices, which are more general than the doubly-stochastic setting
considered in \cite{xian2021faster, gao2022decentralized, chen2022simple, huang2023near, beznosikov2022decentralized, liu2020decentralized}. 
We also carry out  convergence analysis without assuming the smoothness of \textcolor{black}{the} {\em stochastic} loss gradient.

\item We establish  convergence results under more \textcolor{black}{relaxed conditions}, ensuring convergence for both primal and dual solutions.  \textcolor{black}{Previous} works primarily focused on the convergence of \textcolor{black}{the} primal objective  \cite{xian2021faster, gao2022decentralized, chen2022simple, huang2023near}. We show that the dual solution converges at a rate of 
$\mathcal{O}(\frac{1}{T})$ for our method. {\color{black} Please refer to Table \ref{comparison} for the comparison of our results with existing works.}
\end{itemize}

\noindent \emph{Notations.}
We use lowercase normal font \(x\) to represent deterministic scalars or vectors, and lowercase bold font \(\bd{x}\) for stochastic scalars or vectors. 
We use upper case letters $A$ for matrices and calligraphic letters 
\(\bd{\mathcal{X}}\)
for network quantities.
The symbol \(\mathcal{N}_{k}\) denotes the set of neighbors of agent \(k\), and \(a_{\ell k}\) represents the scaling weight for information \textcolor{black}{flowing} from agent \(\ell\) to agent \(k\). 
\(\mathbbm{1}_M\) denotes the $M$-dimensional vector of all ones. The identity matrix of size $M$ is denoted by \(I_M\). 
The Kronecker product operator is denoted by \(\otimes\),  \(\|\cdot\|\) stands for the \(\ell_2\)-norm, and $\langle \cdot, \cdot \rangle$ denotes the inner product operator.
\section{Algorithm Description}
\subsection{Centralized Scenario} 
We \textcolor{black}{start with} a 
centralized scenario
where the data processing and model training
are carried out at a fusion center. Specifically, we consider initially the minimax problem 
with $J(x,y) = \mathbb{E} Q(x,y,\bm{\xi})$ formulated at a stand-alone agent or fusion center.
Several 
variants 
of the stochastic OG
\textcolor{black}{method} will be revisited here 
to motivate our new formulation.
It is noted that \textcolor{black}{these variants} were originally proposed in \cite{korpelevich1976extragradient, popov1980modification} in the
deterministic \textcolor{black}{setting}, \textcolor{black}{whereas} 
we consider the stochastic \textcolor{black}{version}.

The extragradient (EG) method
traces back to the seminal work \cite{korpelevich1976extragradient},  which is based on  
a heuristic idea, \textcolor{black}{namely} ``extrapolation".
It proposed to utilize the gradient at the extrapolated point, rather than the gradient at the current point \textcolor{black}{for carrying out the update.}
Doing so
leads to increased stability by alleviating the rotational force. Such a scheme was also \textcolor{black}{applied to} other optimization problems\cite{antonakopoulos2022extra}. 
The stochastic EG (\textbf{S-EG}) was studied in
the works \cite{NEURIPS2022_Tight, beznosikov2022decentralized}.
This algorithm starts from iteration $i = 0$, say, with randomly initialized variables $\overline{\bd{x}}_{0}, \overline{\bd{y}}_{0}$
and recursively performs the following steps:
\begin{subnumcases}{\textbf{S-EG} ~\label{uda_seg}}
\bd{x}_{i}
= \overline{\bd{x}}_{i} - \mu_x
\nabla_x Q(\overline{\bd{x}}_{i}, \overline{\bd{y}}_{i};
\overline{\bd{\xi}}_{x, i}) \quad   \label{4a} \tag{4a} \\
\bd{y}_{i}
= \overline{\bd{y}}_{i} + \mu_y
\nabla_y Q(\overline{\bd{x}}_{i}, \overline{\bd{y}}_{i};
\overline{\bd{\xi}}_{y,i})\label{4b} \tag{4b} \\
\overline{\bd{x}}_{i+1}
= \overline{\bd{x}}_{i} - \mu_x
\nabla_x Q(\bd{x}^{}_{i}, \bd{y}_{i};
\bd{\xi}_{x,i})\label{4c} \tag{4c} \\
\overline{\bd{y}}_{i+1}
= \overline{\bd{y}}_{i} + \mu_y
\nabla_y Q(\bd{x}_{i}, \bd{y}_{i};
\bd{\xi}_{y, i})\label{4d} \tag{4d}
\end{subnumcases}
Here, 
 $(\bd{x}_{i}, \bd{y}_{i})$ denotes the extrapolated point, and 
$(\overline{\bd{x}}_{i+1},
\overline{\bd{y}}_{i+1})$ is the output. The variables
$(\overline{\bd{\xi}}_{x, i}, 
\overline{\bd{\xi}}_{y, i},
\bd{\xi}_{x, i}, \bd{\xi}_{y, i})
$
are iid samples drawn
 during different steps and for updating different variables. 
We use the subscript and bar notation in these iid samples 
to highlight at which point and
for which variables they are drawn.
For instance,  $\overline{\bd{\xi}}_{x, i}$
denotes an iid sample drawn at iteration $i$ for 
constructing the stochastic gradient at  point $(\overline{\boldsymbol{x}}_i, \overline{\boldsymbol{y}}_i)$ and for updating the
primal variable $\boldsymbol{x}$.
Additionally, \textcolor{black}{the scalars}
 $\mu_x, \mu_y >0$ are the stepsizes or learning  rates.
 The \textbf{S-EG} needs to query
two stochastic gradient oracles to update {\em each} variable. It is also executed in a sequential manner, i.e., determining the extrapolated point first before performing the update.


To reduce the computational complexity,
a single-call version of EG known as past EG (\textbf{PEG}), was introduced in the work \cite{popov1980modification}.
It proposed to 
use the past gradient 
at the point 
$(\bd{x}_{i-1},
\bd{y}_{i-1})$. 
The stochastic form of \textbf{PEG} is characterized by the following steps \cite{hsieh2019convergence, liu2020decentralized}:
\begin{subnumcases}
{\textbf{S-PEG}}
\bd{x}_{i}
= \overline{\bd{x}}_{i} - \mu_x
\nabla_x Q(\bd{x}_{i-1}, \bd{y}_{i-1};
\bd{\xi}_{x, i-1})  \tag{5a} \label{eq5a}\\
\bd{y}_{i}
= \overline{\bd{y}}_{i} + \mu_y
\nabla_y Q(\bd{x}_{i-1}, \bd{y}_{i-1};
\bd{\xi}_{y, i-1}) \tag{5b}\label{eq5b} \\
\overline{\bd{x}}_{i+1}
= \overline{\bd{x}}_{i} - \mu_x
\nabla_x Q(\bd{x}_{i}, \bd{y}_{i};
\bd{\xi}_{x, i}) \tag{5c}\label{eq5c} \\
\overline{\bd{y}}_{i+1}
= \overline{\bd{y}}_{i} + \mu_y
\nabla_y Q(\bd{x}_{i}, \bd{y}_{i};
\bd{\xi}_{y,i}) \tag{5d}\label{eq5d} 
\end{subnumcases}
\textbf{PEG} 
can also be derived from the
forward-reflected-backward framework \cite{malitsky2020forward}. It has attracted wide interest 
in theoretical aspects 
and obtains empirical success in practical applications
\cite{gorbunov2022last, daskalakis2018training, liu2020decentralized}.

The standard form of \textbf{S-PEG},  also referred to as the stochastic OG (\textbf{S-OG}), is often utilized for theoretical analysis \cite{NEURIPS2022_Tight}.
\textcolor{black}{The updates}  in Eqs. \eqref{eq5a}--(5d) can be simplified by rewriting them in terms of 
 $(\bd{x}_{i+1}, \bd{y}_{i+1})$
at each iteration $i$ \cite{NEURIPS2022_Tight, hsieh2019convergence}:
\begin{subnumcases}
{\textbf{S-OG}}
\bd{x}_{i+1}
= \bd{x}_{i} - \mu_x\Big[2
\nabla_x Q(\bd{x}_{i}, \bd{y}_{i};
\bd{\xi}_{x, i})  \notag \\
\qquad \quad  +  
\nabla_x Q(\bd{x}_{i-1}, \bd{y}_{i-1};
\bd{\xi}_{x, i-1}) \Big] \tag{6a} \label{eq6a} \\
\bd{y}_{i+1}
= \bd{y}_{i} + \mu_y \Big[2
\nabla_y Q(\bd{x}_{i}, \bd{y}_{i};
\bd{\xi}_{y, i}) \notag \\ \qquad \quad  - \nabla_y Q(\bd{x}_{i-1}, \bd{y}_{i-1};
\bd{\xi}_{y, i-1}) \Big]\tag{6b} \label{eq6b}
\end{subnumcases}
\textbf{S-OG} involves
two different samples, namely 
\textcolor{black}{$\bm{\xi}_{x,i}$ and $\bm{\xi}_{x,i-1}$},
for updating the primal variable in different steps (similar to the dual variable).
We observe that $\bd{\xi}_{x, i}$ 
and $\bd{\xi}_{x, i-1}$
are actually approximating different losses whose landscape differs. The gradient
$\nabla_x Q(\bd{x}_{i}, \bd{y}_{i};
\bd{\xi}_{x, i})$ constructed by $\bd{\xi}_{x, i}$
can  deviate
from that
at the same point $(\bd{x}_{i}, \bd{y}_{i})$ when it is 
evaluated at $\bd{\xi}_{x, i-1}$.
In other words, \textbf{S-OG} is not making the ``optimistic" direction under the same landscape in a strict sense.
When the sample variance is high,
the deviation between the loss landscapes constructed
from $\bd{\xi}_{x, i}$ and $\bd{\xi}_{x, i-1}$
could be  high,  leading to a poor approximation of the update direction \cite{mishchenko2020revisiting}. On the other hand,
the stochastic gradients involved in \eqref{eq6a}--(6b) are not martingale processes, which cause difficulty for analysis (see Section III-E for discussion).

\textcolor{black}{We instead} introduce  
 a new formulation using the {\em same-sample} approximation. For each variable, it
  utilizes the gradients from the same loss landscape to perform
 ``optimistic'' scheme; these losses are also unbiased for the true risk. This can be verified by taking expectation over the distribution of the current samples $(\bd{\xi}_{x, i}$, $\bd{\xi}_{y, i})$.
This approach involves recomputing the stochastic gradient at the point  $(\bd{x}_{i-1},\bd{y}_{i-1})$
using the current samples $(\bd{\xi}_{x, i}, \bd{\xi}_{y, i})$:
\begin{subnumcases}
{\textbf{SS-OG}} 
  \bd{x}_{i+1}
= \bd{x}_{i} - 2\mu_x\Big[
\nabla_x Q(\bd{x}_{i}, \bd{y}_{i};
\bd{\xi}_{x, i})  \notag  \\
\qquad \quad  +  
\nabla_x Q(\bd{x}_{i-1}, \bd{y}_{i-1};
\bd{\xi}_{x, i}) \Big]  \tag{7a} \label{eq7a} \\
\bd{y}_{i+1}
= \bd{y}_{i} + 2\mu_y \Big[
\nabla_y Q(\bd{x}_{i}, \bd{y}_{i};
\bd{\xi}_{y, i}) \notag \\ \qquad \quad-  \nabla_y Q(\bd{x}_{i-1}, \bd{y}_{i-1};
\bd{\xi}_{y, i}) \Big] \tag{7b}\label{eq7b} 
\end{subnumcases}
\textbf{SS-OG}
requires double call
of stochastic gradient oracles for {\em each} variable. However,
it can be executed in
parallel with some extra memory overhead, making it preferred \textcolor{black}{in comparison} to the sequential approaches, e.g., \textbf{S-EG} and \textbf{S-PEG}. 


\begin{algorithm}[t]
\caption{\textbf{S}tochastic \textbf{S}ame-Sample \textbf{O}ptimistic \textbf{G}radient (\textbf{SS-OG})}
\label{example-algorithm-single}
\begin{algorithmic}[1]
\footnotesize
\renewcommand{\REQUIRE}{\item[\textbf{Initialize:}]}
\REQUIRE $\boldsymbol{x}_{i}, \boldsymbol{y}_{i} { \scriptstyle (i= -1, -2)}$, step {\color{black} sizes} $\mu_x$, $\mu_y$.
\FOR{ $i = 0,1, \dots$}    
\STATE \underline{\text{Compute  stochastic gradient with  samples} $\boldsymbol{\xi}_{x, i}, \boldsymbol{\xi}_{y, i}$}
\STATE  
 $\displaystyle 
 \begin{aligned}
 {\boldsymbol{g}}_{x,i-1} =    
 2\nabla_{x} Q(\boldsymbol{x}_{i-1}, \boldsymbol{y}_{ i-1} ;\boldsymbol{\xi}_{x, i})-  \nabla_{x} Q(\boldsymbol{x}_{ i-2}, \boldsymbol{y}_{ i-2} ;\boldsymbol{\xi}_{x, i})
 \end{aligned}
 $
\STATE    $\displaystyle \begin{aligned}
 {\boldsymbol{g}}_{y,i-1} = 
2\nabla_{y} Q(\boldsymbol{x}_{i-1}, \boldsymbol{y}_{ i-1} ;\boldsymbol{\xi}_{y,i})- \nabla_{y} Q(\boldsymbol{x}_{ i-2}, \boldsymbol{y}_{ i-2} ;\boldsymbol{\xi}_{y,i})
\end{aligned}$
\STATE  \underline{\text{Adaptation}}
\STATE$\displaystyle  {\boldsymbol{x}}_{i}  = \boldsymbol{x}_{i-1} 
     - \mu_x \boldsymbol{g}_{x,i-1}$ 
\STATE
     $ \displaystyle  {\boldsymbol{y}}_{i}  = \boldsymbol{y}_{i-1} 
     + \mu_y \boldsymbol{g}_{y,i-1}$  \\
\ENDFOR
\end{algorithmic}
\end{algorithm}
\subsection{Distributed scenario}
In this \textcolor{black}{section},
we \textcolor{black}{extend} \textbf{SS-OG}
\textcolor{black}{to the {\em distributed} setting \eqref{risk}--\eqref{local_risk}}.
Under this scenario,
each agent $k$ utilizes
 a local sample $\{\boldsymbol{\xi}_{k}\}$
to approximate the true gradient by using its loss value. \textcolor{black}{
Several distributed learning strategies for minimization problems have been investigated in \cite{sayed2014adaptation,sayed2022inference} and the references therein.} 
\textcolor{black}{Here, we} integrate  \eqref{eq6a}-(6b)
with \textcolor{black}{the} adapt-then-combine (ATC) diffusion learning 
\textcolor{black}{strategy from \cite{sayed2014adaptation,sayed2022inference}.  This strategy has been shown in these {\color{black} references} to outperform the consensus strategy, which has been applied to minimax problems as well in works such as \cite{liu2020decentralized,gao2022decentralized}. For this reason, we focus on diffusion learning and refer to the proposed algorithm as the}
\textbf{D}iffusion \textbf{S}tochastic \textbf{S}ame-Sample \textbf{O}ptimistic \textbf{G}radient (\textbf{DSS-OG}). \textcolor{black}{It is listed in}
\textbf{Algorithm 1}. {\color{black} In this algorithm, $a_{\ell k}$ is the scalar weight used by agent $k$ to scale information received from agent $\ell \in \mathcal{N}_k$ and $a_{\ell k}=0$ for $\ell \notin \mathcal{N}_k$.}
This algorithm starts from $i=0$ with 
randomly initialized local
weights $\boldsymbol{x}_{k,i}, \boldsymbol{y}_{k,i} (i=-1,-2)$. 
Each agent  first collect two iid samples, namely $\boldsymbol{\xi}^{x}_{k,i}$ and $ \boldsymbol{\xi}^{y}_{ k,i}$, to approximate the OG direction for updating the local primal and dual variables, respectively. 
After then the local models of each agent are sent to its immediate neighbors 
for  aggregation {\color{black} using the weights $\{a_{\ell k}\}$.}

\begin{algorithm}[t]
\caption{\textbf{D}iffusion \textbf{S}tochastic \textbf{S}ame-Sample \textbf{O}ptimistic \textbf{G}radient (\textbf{DSS-OG})}
\label{example-algorithm}
\begin{algorithmic}[1]
\footnotesize
\renewcommand{\REQUIRE}{\item[\textbf{Initialize:}]}
\REQUIRE $\boldsymbol{x}_{k,i}, \boldsymbol{y}_{k,i} { \scriptstyle (i= -1, -2)}$, step {\color{black} sizes} $\mu_x$, $\mu_y$.
\FOR{ $i = 0,1, \dots$}    
\FOR{each agent $k$}
\STATE \underline{\text{Compute  stochastic gradient with  sample} $\boldsymbol{\xi}^{x}_{k,i}, \boldsymbol{\xi}^{y}_{ k,i}$}
\STATE  
 $\displaystyle 
 \begin{aligned}
 {\boldsymbol{g}}_{x,k,i-1} = &
\ 2\nabla_{x} Q_k(\boldsymbol{x}_{k, i-1}, \boldsymbol{y}_{k, i-1} ;\boldsymbol{\xi}^{x}_{k, i}) \\   
 &-  \nabla_{x} Q_k(\boldsymbol{x}_{k, i-2}, \boldsymbol{y}_{k, i-2} ;\boldsymbol{\xi}^{x}_{ k, i})
 \end{aligned}
 $
\STATE    $\displaystyle \begin{aligned}
 {\boldsymbol{g}}_{y,k,i-1} = &  \
2\nabla_{y} Q_k(\boldsymbol{x}_{k, i-1}, \boldsymbol{y}_{k, i-1} ;\boldsymbol{\xi}^{y}_{k, i}) \\
&- \nabla_{y} Q_k(\boldsymbol{x}_{k, i-2}, \boldsymbol{y}_{k, i-2} ;\boldsymbol{\xi}^{y}_{k, i})
\end{aligned}$
\STATE  \underline{\text{Adaptation}}
\STATE$\displaystyle  {\boldsymbol{\phi}}_{k,i}  = \boldsymbol{x}_{k,i-1} 
     - \mu_x \boldsymbol{g}_{x,k,i-1}$ 
\STATE
     $ \displaystyle  {\boldsymbol{\psi}}_{k,i}  = \boldsymbol{y}_{k,i-1} 
     + \mu_y \boldsymbol{g}_{y,k,i-1}$ 
\STATE \underline{\text{Combination}}
\STATE  $\boldsymbol{x}_{k,i} = \underset{\ell \in \mathcal{N}_k}{\sum} a_{\ell k} 
{\boldsymbol{\phi}}_{\ell,i}, \quad \displaystyle  \boldsymbol{y}_{k,i} = \underset{\ell \in \mathcal{N}_k}{\sum} a_{\ell k} 
{\boldsymbol{\psi}}_{\ell,i}$ \\
\ENDFOR
\ENDFOR
\end{algorithmic}
\end{algorithm}

For convenience of \textcolor{black}{the} analysis, we introduce 
the 
following network quantities:
\vspace{-1em}
\begin{subequations}
    \begin{align} \hspace{3em}{\boldsymbol{x}}_{c, i} &\triangleq 
      \sum_{k=1}^{K} p_k \boldsymbol{x}_{k,i} 
      \in \mathbb{R}^{M_1 \times 1} \quad \text{(centroid)}
    \\ 
\boldsymbol{g}_{x,c,i} 
    &\triangleq \sum_{k=1}^K p_k {\boldsymbol{g}}_{x,k,i}  \in \mathbb{R}^{M_1 \times 1}  
    \\
      \boldsymbol{\mathcal{X}}_{i} &\triangleq  \mbox{\rm col}\{\boldsymbol{x}_{1,i}, \dots, \bm{x}_{K,i}\}
     \in  \mathbb{R}^{M_1K \times 1}
     \\     \boldsymbol{{\mathcal{X}}}_{c, i} &\triangleq  
\mbox{\rm col}\{
       \boldsymbol{x}_{c,i}, \dots, 
 \boldsymbol{x}_{c,i}\} \in \mathbb{R}^{M_1K \times 1}
 \\
 \bm{\mathcal{G}}_{x, i}
 &\triangleq 
 \mbox{\rm col}
 \{\bm{g}_{x,1,i}, \dots, \bm{g}_{x,K,i}\} \in \mathbb{R}^{M_1K \times 1}
\end{align}
\end{subequations}
and similarly for ${\boldsymbol{y}}_{c, i}, \boldsymbol{g}_{y,c,i} \in \mathbb{R}^{M_2 \times 1 }, \boldsymbol{\mathcal{Y}}_{i},
\boldsymbol{\mathcal{Y}}_{c, i}, \bm{\mathcal{G}}_{y, i}\in \mathbb{R}^{M_2K \times 1}$.
Additionally, 
we use ${{g}}_{x,c,i},{{g}}_{y,c,i},{\mathcal{G}}_{x, i},{\mathcal{G}}_{y, i}$
\textcolor{black}{refer to deterministic realizations of the random quantities} $\boldsymbol{g}_{x,c,i}$,  $\boldsymbol{g}_{y,c,i}$, $\bm{\mathcal{G}}_{x, i}$, $\bm{\mathcal{G}}_{y, i}$.
\textcolor{black}{If we further introduce the network combination matrices:  
\begin{subequations}
    \begin{align}
                A&=[a_{\ell k}] \in \mathbb{R}^{K \times K} \\
\mathcal{A}_1 &\triangleq A \otimes I_{M_1}\in \mathbb{R}^{M_1K \times M_1K} \\ \mathcal{A}_2 &\triangleq A \otimes I_{M_2}\in \mathbb{R}^{M_2K \times M_2K}
    \end{align}
\end{subequations}
then, the} 
\textbf{DSS-OG} \textcolor{black}{recursions}  be expressed \textcolor{black}{globally} the \textcolor{black}{following} recursion:
\textcolor{black}{
\begin{subequations}
\begin{align}
\label{networkupdate}
\boldsymbol{\mathcal{X}}_i &= 
   \mathcal{A}_1^\top \{ \boldsymbol{\mathcal{X}}_{i-1} - \mu_x
    \bm{\mathcal{G}}_{x, i-1}\} \\
\boldsymbol{\mathcal{Y}}_i &= 
    \mathcal{A}_2^\top\{  \boldsymbol{\mathcal{Y}}_{i-1} + \mu_y
     \bm{\mathcal{G}}_{y, i-1} \}   
\end{align}  
\end{subequations}
}
\setcounter{equation}{10}
\vspace{-2em}
\section{Convergence Analysis}
We now proceed to
\textcolor{black}{analyze the convergence properties of} 
  \textbf{DSS-OG}.
\textcolor{black}{We do so under the following assumptions, which are generally more relaxed than similar conditions in the prior literature. }

\vspace{-1em}
\subsection{Assumptions}
\begin{Assumption}\label{Lipschitz}
We assume the  {\color{black}gradients} associated with each local {\color{black}risk} function \textcolor{black}{are} $L_f$-Lipschitz, i.e.,
\begin{align} 
& \| \nabla_{w} J_k (x_1, y_1) - \nabla_w J_k (x_2, y_2)\| \quad (w = x \text{ or }y) \\
 \quad &\le   
L_f\Big(\|x_1 - x_2\|  
 + \|y_1 - y_2\|\Big)  \notag 
\end{align}
\end{Assumption}
\begin{remark}
In this work,
we will only adopt the above assumption for true gradients. 
{\color{black}This} is in  contrast to the works  \cite{xian2021faster, gao2022decentralized, chen2022simple,huang2023near},
where smoothness of the local stochastic loss 
\textcolor{black}{is assumed instead}.
 It should be noted that if we only assume Assumption \ref{Lipschitz},
 the computational benefits of all momentum techniques
used in \cite{xian2021faster, huang2023near}
 are nullified,
with the gradient evaluation complexity \textcolor{black}{degrading} 
to $\mathcal{O}(\varepsilon^{-4})$.
This is the tight lower bound \textcolor{black}{for} solving a nonconvex problems developed in \cite{arjevani2023lower}. Moreover, the primal objective introduced in \eqref{primalobjective} is essentially nonconvex.
$\hfill{\square}$
\end{remark}
 
 Since the primal problem is to minimize \textcolor{black}{over} $x$ \textcolor{black}{in} \eqref{risk}, 
we introduce \textcolor{black}{the} primal objective\cite{xian2021faster, gao2022decentralized, chen2022simple,huang2023near}: 
\begin{equation} \label{primalobjective}
    P(x) = {\max}_y  \ J(x, y)
\end{equation}
\begin{Assumption} \label{boundPhi}
We assume  $P(x)$ is lower bounded, i.e.,
$P^\star =\operatorname{inf}_{x} P(x) > -\infty$.
\end{Assumption}

\begin{Assumption} \label{Expectation}
{\color{black}We denote} the filtration generated by the random {\color{black}processes as}
$\boldsymbol{\mathcal{F}}_{i} = \{(\bd{x}_{k,j},\bd{y}_{k,j}) \mid k=1, \dots, K,
j =-2, -1, \dots, i \}$.
We assume
the  stochastic {\color{black}gradients  are} unbiased {\color{black} with} bounded variance while taking expectation {\color{black} over the random sample $\bd{\xi}^{w}_{k, i}$} conditioned
on
$\boldsymbol{\mathcal{F}}_{i}$, i.e.,
    \begin{equation}
    \begin{aligned}
 &  \mathbb{E}\Big\{ \nabla_{w} Q_k(\bd{x}_{k,i}, \bd{y}_{k,i};\boldsymbol{\xi}^{w}_{k, i}) \mid       \boldsymbol{\mathcal{F}}_{i}\Big\} = \nabla_{w} J_k(\bd{x}_{k,i},\bd{y}_{k,i}) \\
&      \mathbb{E}
      \Big\{\|\nabla_{w} Q_k(\bd{x}_{k, i}, \bd{y}_{k, i};\boldsymbol{\xi}^{w}_{k,i}) -  \nabla_{w} J_k(\bd{x}_{k, i}, \bd{y}_{k, i})\|^2 \mid \boldsymbol{\mathcal{F}}_{i} \Big\} \\
         & \le {\sigma^2_k}  \quad (w = x \text{ or }y)
    \end{aligned}
    \end{equation} Moreover,
    we assume the data samples 
    $\boldsymbol{\xi}^{w}_{k, i}$
    are independent of each other 
    for all  $k, i, w$.
    \end{Assumption}
\begin{Assumption} \label{boundnetwork}
    We assume the disagreement between the local 
    and global \textcolor{black}{gradients} is bounded, i.e.,
    \begin{equation}
    \begin{aligned}
         \|\nabla_{w}  J_k(\bd{x}_{k,i}, \bd{y}_{k, i}) -
         \nabla_{w}  J(\bd{x}_{k,i}, \bd{y}_{k,i})\|^2 &\le G^2  \ (w = x \text{ or }y)
    \end{aligned} 
    \end{equation}
\end{Assumption}
The above assumptions can  be relaxed 
by adopting more sophisticated distributed learning strategies, e.g., 
exact diffusion and gradient tracking \cite{yuan2018exact,alghunaim2022unified,alghunaim2020decentralized}; 
we leave \textcolor{black}{these extensions} for future works.
\begin{Assumption} \label{LeftStochastic}
 The graph is strongly connected and the matrix $A = [a_{\ell, k}]$
is left-stochastic.
As a result, it holds that 
$\mathbbm{1}^\top A = \mathbbm{1}^\top$
and $A^r$ has strictly positive entries for some integer $r$.
\end{Assumption}
This assumption ensures that $A$
has a single maximum eigenvalue 
{\color{black} at $1$ with}
Perron eigenvector 
$p$ \textcolor{black}{with positive entries such } that $Ap = p, \mathbbm{1}^\top p =1$.
 Furthermore, the combination matrix $A$
   admits the Jordan decomposition
 $A = V J V^{-1}$  {\color{black}where} \cite{sayed2014adaptation}:
\begin{equation} \label{Jordan}
V
= \begin{bmatrix}
    p, V_R
\end{bmatrix}, J = \begin{bmatrix}
1 & 0\\
0& J_\gamma
\end{bmatrix},
V^{-1}
=
\begin{bmatrix} 
\mathbbm{1}^\top \\
V^\top_L
\end{bmatrix}
\end{equation}
{\color{black}Here, 
the submatrix $J_\gamma$
consists of Jordan blocks 
with arbitrarily small values on the lower diagonal}, and
\textcolor{black}{it holds that} $\mathbbm{1}^\top V_R = 0$, $V^\top_L V_R = I_{K-1}$, 
$\|J_\gamma\| <1$.
For  convenience of the analysis,
we {\color{black} let} 
$
 \mathcal{J}_{\gamma_1} \triangleq J_\gamma \otimes I_{M_1},   \mathcal{J}_{\gamma_2} \triangleq J_\gamma \otimes I_{M_2},
  \mathcal{V}_{R_1} \triangleq V_{R} \otimes I_{M_1}, \mathcal{V}_{R_2} \triangleq
  V_{R} \otimes I_{M_2},
   \mathcal{V}_{L_1} \triangleq V_L \otimes I_{M_1},
    \mathcal{V}_{L_2} \triangleq V_L \otimes I_{M_2}$.
    It is not difficult to verify
    $\|\mathcal{J}_{\gamma_1}\| = \|\mathcal{J}_{\gamma_2}\|, \|\mathcal{V}_{R_1} \| = \|\mathcal{V}_{R_2}\|,
   \|V_{L_1}\| = \|V_{L_2}\| $.



    


\vspace{-1em}
\subsection{Main Results}
In Theorem \ref{maintheorem}, we prove the  convergence rate
of the best-iterate
of  \textcolor{black}{the} primal objective, which is the standard criterion for convergence analysis involving the nonconvex problem. In Theorem \ref{maintheorem2}, we show the last-itearate convergence of the  dual optimality gap.
Given the conditions
from these theorems, 
we establish \textcolor{black}{the} complexity 
of the \textbf{DSS-OG} in finding an $\varepsilon$-stationary point.

\setcounter{Theorem}{0}
\begin{Theorem} \label{maintheorem}
Under Assumptions
\ref{muPL}--\ref{LeftStochastic},
choosing  step sizes 
\begin{equation}
    \scalebox{1.0}{$\mu_y = \min \{ \frac{1}{\nu}, \frac{2L^2_f}{\nu \beta_1}, \frac{1 -\|\mathcal{J}^\top_{\gamma_1}\|^2}{T^{1/2}},1\}$}
\end{equation}
\begin{equation}
    \scalebox{1.0}{$\mu_x = \min \{
\frac{\nu^2\mu_y}{6L^2_f}, \frac{1}{2(L+L_f)}, \frac{1}{4L}, \frac{\mu_y}{4\tau_3}, \frac{1}{2\sqrt{\tau_1+\beta_1}}, \mu_y
\}$}
\end{equation}
the non-asymptotic convergence rate
of \textcolor{black}{the} primal objective is given by:
\begin{align}
\label{primalconvergence}
\frac{1}{T}\sum_{i=0}^{T-1} \mathbb{E}\|\nabla P(\boldsymbol{x}_{c,i})\|^2
\le  \mathcal{O}\Big(\frac{1}{{T}^{1/2}}\Big) + \mathcal{O}\Big(\frac{1}{T}\Big)  
+ \mathcal{O}\Big(\frac{1}{T^2}\Big)
\end{align}
where  \scalebox{0.9}{$L =L_f + \frac{L^2_f}{\nu}$} is the Lipschitz constant associated with $P(x)$ and 
\begin{equation}
\scalebox{1.0}{$
\tau_1 \triangleq \frac{80L^4_f}{\nu^2} ,
\tau_3 \triangleq\frac{80L^2_f}{\nu^2}, \beta_1 \triangleq \frac{96KL_f^4}{\nu^2} + 48KL^2_f$}
\end{equation}
are constants.
\textcolor{black}{Furthermore}, \textbf{DSS-OG} outputs \textcolor{black}{a} primal $\varepsilon$-stationary  point after $T_0 = \mathcal{O}(\varepsilon^{-4})$  iterations 
and gradient evaluation
complexity, i.e., 
\begin{align}
\label{primalsolution1}
\mathbb{E}\|\nabla P({\boldsymbol{x}}^\star_{c, T_0})\|^2 &= 
     \inf_{i=0,\dots, T_0-1}
    \mathbb{E}\|\nabla P(\boldsymbol{x}_{c,i})\|^2 \\
    &\le 
    \frac{1}{T_0} \sum_{i=0}^{T_0-1}
  \mathbb{E}\|\nabla P(\boldsymbol{x}_{c,i})\|^2
  \le \mathcal{O}(\varepsilon^{2}) \notag
\end{align}
See Appendix \ref{proofoftheorem1} for proof.
\end{Theorem}
Theorem \ref{maintheorem}
states that 
by adopting a two-time-scale scheme for the step sizes
$\mu_x$ and $\mu_y$, \textbf{DSS-OG}
is able to output
an $\epsilon$-stationary point for the primal objective 
with
a sublinear rate 
dominated by 
$\mc{O}\Big(\frac{1}{T^{1/2}}\Big)$.
The two-time-scale condition reflects 
the unsymmetric assumption 
of the risk function  over the primal and dual variables.
Similar conditions 
were also established 
in the works 
\cite{NEURIPS2022_Tight,lin2020gradient,yang2022faster} when asymmetric conditions are assumed over the primal and dual variables.

\begin{Theorem} \label{maintheorem2}
Under Assumptions
\ref{muPL}--\ref{LeftStochastic},
and after iteration numbers $T_0$,
choosing step size 
\begin{equation}
    \scalebox{1.0}{$\mu_x =0, \quad \mu_y = \min\{\sqrt{\frac{\nu}{12KL^2_f}}, \frac{2T+1}{\nu(T+1)^2}\}$}
\end{equation}
the non-asymptotic bound of the dual optimality gap 
regarding the last-iterate
is given by:
\begin{align}
\label{dualconvergence}
 \mathbb{E}[P({\boldsymbol{x}}^\star_{c,T_0})
- 
J({\boldsymbol{x}}^\star_{c,T_0}, \boldsymbol{y}_{c,T}) ] 
 \le \mathcal{O}\Big(\frac{1}{T}\Big) + \mathcal{O}\Big(\frac{1}{T^2}\Big)
\end{align}
That is, \textbf{DSS-OG} outputs a dual $\varepsilon$-stationary point after $T_1= \mathcal{O}(\varepsilon^{-2})$  iterations 
and gradient evaluation
complexity, i.e., 
\begin{equation} \label{condition}
\begin{aligned}  
&\mathbb{E}[
P({\boldsymbol{x}}^\star_{c,T_0})
    -    J({\boldsymbol{x}}^\star_{c,T_0}, {\boldsymbol{y}}^\star_{c, T_0+T_1}) ]  \le \mathcal{O}(\varepsilon^{2})
\end{aligned}
\end{equation}
See Appendix \ref{proofoftheorem2} for proof.
\end{Theorem}
The above theorem 
implies that 
finding a dual solution that shrinks the
dual optimality gap 
has a faster sublinear convergence rate of $\mc{O}\Big(\frac{1}{T}\Big)$ than the primal convergence rate.
 Eqs. \eqref{primalconvergence} and \eqref{dualconvergence} 
 also show the fundamental difference of the complexity in finding the solution for a nonconvex problem and  for a PL problem.
From this theoretical aspect,
 it is  hard to guarantee 
 a pair of primal and dual solutions with the same accuracy level \eqref{29a}--\eqref{29b} simultaneously.
Fortunately, 
we can prove that the same accuracy level of  the primal and dual solutions can be obtained in an asynchronous manner (See Theorem \ref{maintheorem3}).
\begin{Theorem} \label{maintheorem3}
After iterations $T_0 +T_1 = \mathcal{O}(\varepsilon^{-4}) +\mathcal{O}(\varepsilon^{-2})$ of running \textbf{DSS-OG},
we can find  the network centroid 
$({\boldsymbol{x}}^\star_{c,T_0}, {\boldsymbol{y}}^\star_{c,T_0+T_1} )$
 that
satisfies the following  first-order stationary conditions:
\begin{subequations}
\begin{align} 
\mathbb{E}\|\nabla_{x} J({\boldsymbol{x}}^\star_{c,T_0}, {\boldsymbol{y}}^\star_{c,T_0+T_1})\|^2
& \le \varepsilon^2
\label{29a} 
\\
\mathbb{E}\| \nabla_{y} J({\boldsymbol{x}}^\star_{c,T_0}, {\boldsymbol{y}}^\star_{c,T_0+T_1})\|^2
& \le \varepsilon^2
\label{29b} 
\end{align}
\end{subequations}
for an arbitrarily small constant $\varepsilon$.
The solutions points 
$\boldsymbol{x}^\star_{c,T_0}, {\boldsymbol{y}}^\star_{c,T_0+T_1}$
are found in  different time instance where
 the subscripts in $({\boldsymbol{x}}^\star_{c,T_0}, {\boldsymbol{y}}^\star_{c,T_0+T_1})$ indicate the exact iterations when they are found.
See Appendix \ref{proofoftheorem3} for proof.
\end{Theorem}

\subsection{Comparison between \textbf{S-OG} and  \textbf{SS-OG}}

We point out the key differences between \textbf{SS-OG} and \textbf{S-OG}:
\begin{itemize}
    \item The stochastic gradients involved in \textbf{S-OG} are not {\em martingale} process. To see this, we observe that the backward gradients
    $[\nabla_x Q(\bd{x}_{i-1}; \bd{y}_{i-1};
\bd{\xi}_{x, i-1}), 
    \nabla_y Q(\bd{x}_{i-1}, \bd{y}_{i-1};
\bd{\xi}_{y, i-1})]$ used at iteration $i$ are 
    constructed by the past samples $(\bd{\xi}_{x, i-1}, \bd{\xi}_{y, i-1})$. They are biased estimators when taking expectation over the current samples 
    $(\boldsymbol{\xi}_{x,i}, \boldsymbol{\xi}_{y,i})$. 
    In contrast, it is easy to verify the stochastic gradients involved in \textbf{SS-OG} are unbiased.
    \item \textbf{SS-OG} strictly performs 
    the ``optimistic" scheme under the same loss landscape. 
    The losses utilized there are 
     unbiased estimators for the local risk. In contrast,
    \textbf{S-OG}
    involves 
    gradients from two different loss landscape. 
    The
gradient direction at the same point in two different loss
landscapes can be notably different when data variance is high.
    \item 
    We will bound the  terms involving
     $\mathbb{E}\|\nabla_x J(\boldsymbol{x}_{c,i}, \boldsymbol{y}_{c,i}) - {g}_{x,c,i}\|^2$
     for \textbf{DSS-OG}
     rather than $\mathbb{E}\|\nabla_x J(\boldsymbol{x}_{c,i}, \boldsymbol{y}_{c,i}) - \boldsymbol{g}_{x,c,i}\|^2$ for  distributed \textbf{S-OG}.
     To see this,
    we can repeat the arguments in \eqref{startsmooth1} and use the fact that the two terms involved in the inner product $\langle 
  \nabla {{P}}(\boldsymbol{x}_{c,i}),
  \boldsymbol{g}_{x,c,i}
  \rangle$ are correlated for distributed \textbf{S-OG}. 
Therefore, we observe that the
      distributed \textbf{S-OG} needs to bound
     the deviation 
     between the deterministic and 
    stochastic quantities. 
  The consequence of such deviation is that a constant on the order of $\mathcal{O}(\sigma^2)$ will appear in the final bound because the 
  Lipschitz continuity to be used is only assumed for determistic quantities. 
  In
   constrast,  we only need to bound the deviation between deterministic quantities for \textbf{DSS-OG}. Therefore, employing a large batch size will become necessary for \textbf{S-OG} in order to control the magnitude of $\mathcal{O}(\sigma^2)$.
\end{itemize}

\vspace{-1
em}
\section{Computer simulation}
\begin{figure*}[!htbp]
\centering
\subfigure[
\scalebox{0.9}{$\frac{1}{K}\sum_{k=1}^{k}
(\|\nabla_x Q_k(\bd{x}_{k,i}, \bd{y}_{k,i})\| +
\|\nabla_y Q_k(\bd{x}_{k,i}, \bd{y}_{k,i})\|)$}]{\includegraphics[width=0.38\linewidth]{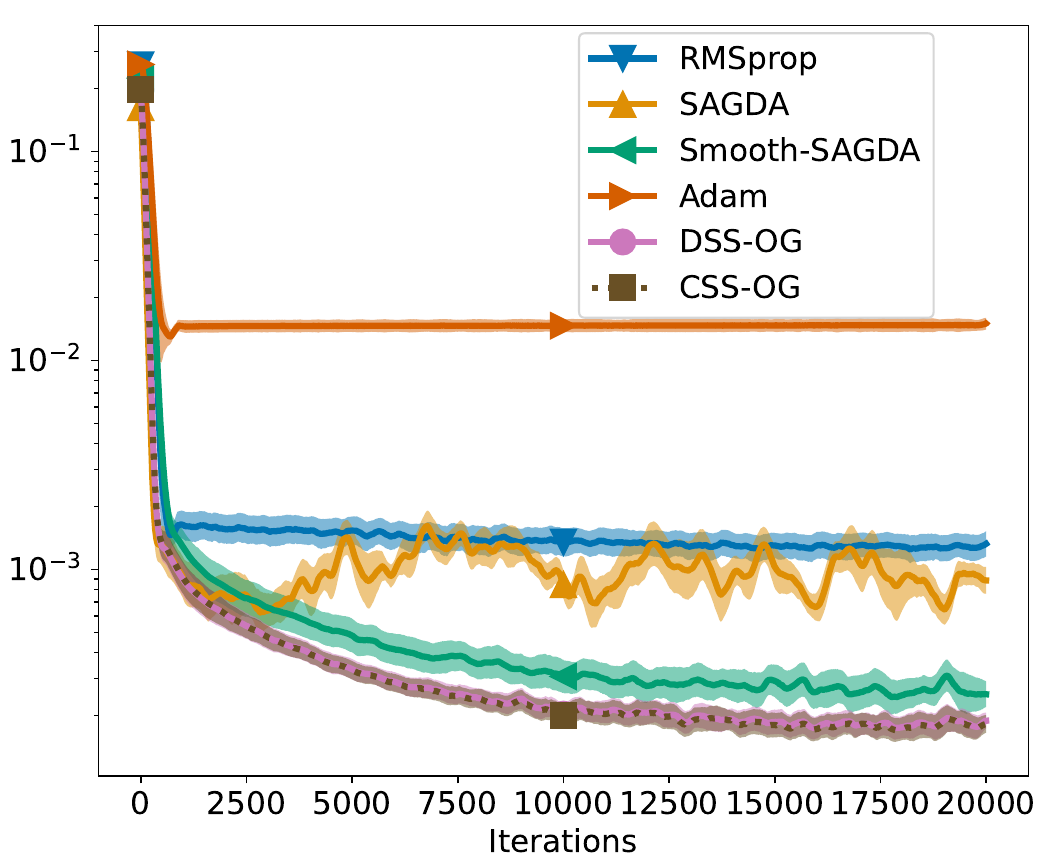}
\label{fig:example1_variance0001_grad}
 }
 \subfigure[\scalebox{0.9}{$\frac{1}{K}\sum_{k=1}^{K}
 (\|\hat{\pi}_{k,i} -\pi_k\|^2+\|\hat{\sigma}_{k,i} - \sigma_k\|^2)$}]{
\includegraphics[width=0.38\linewidth]{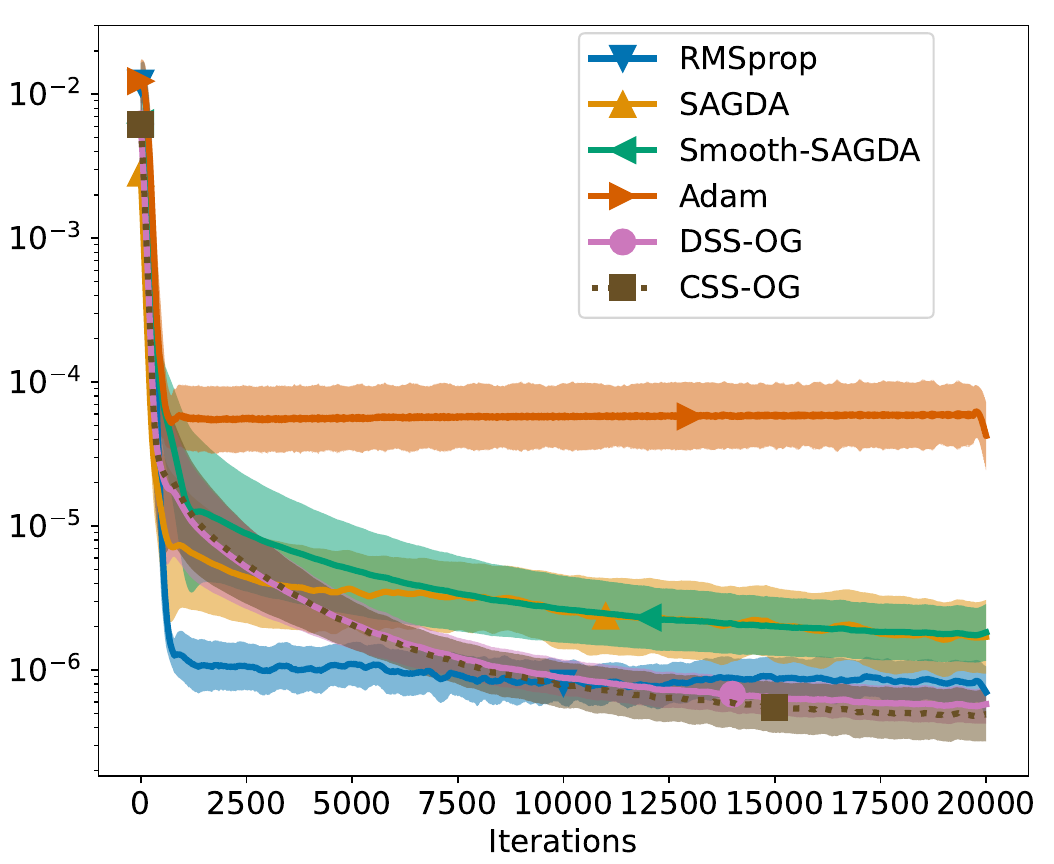}
\label{fig:example1_variance0001_mse}
 }
\subfigure[\scalebox{0.9}{$\frac{1}{K}\sum_{k=1}^{k}
(\|\nabla_x Q_k(\bd{x}_{k,i}, \bd{y}_{k,i})\| +
\|\nabla_y Q_k(\bd{x}_{k,i}, \bd{y}_{k,i})\|)$}]{
\includegraphics[width=0.38\linewidth]{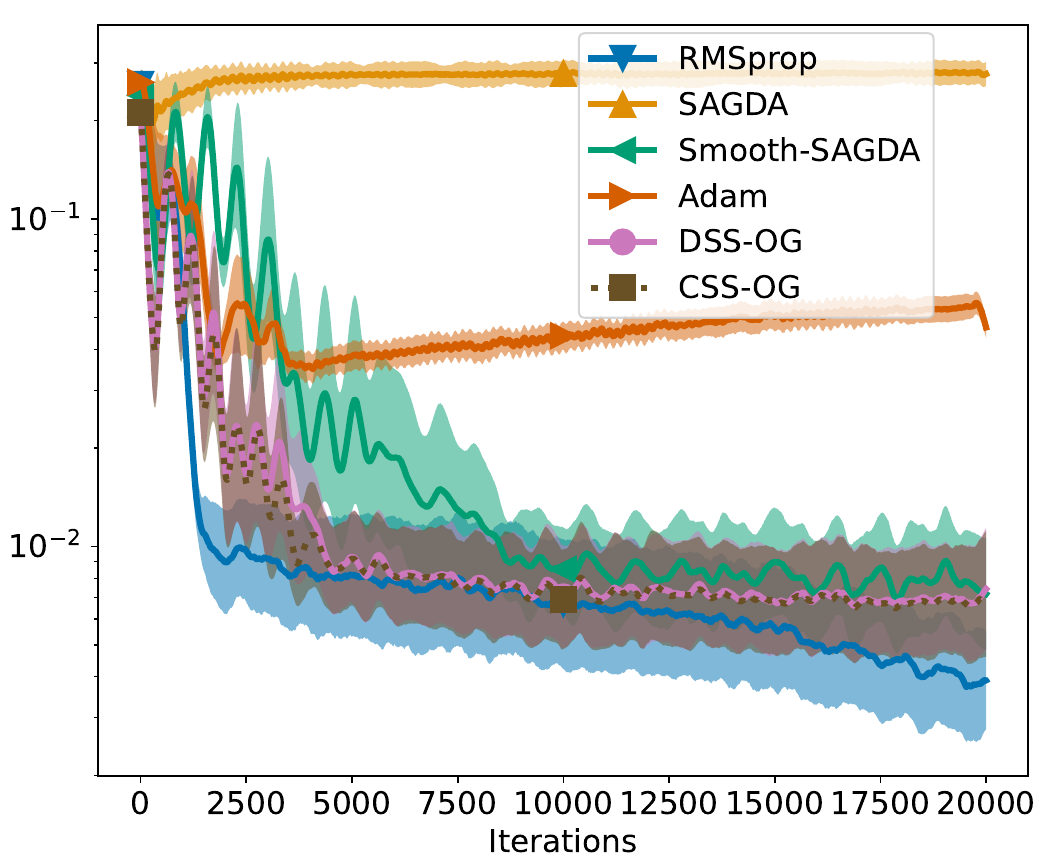} \label{fig:example1_variance01_grad}
 }
\subfigure[\scalebox{0.9}{$\frac{1}{K}\sum_{k=1}^{K}(
 \| \hat{\pi}_{k,i} -\pi_k\|^2+\|\hat{\sigma}_{k,i} - \sigma_k\|^2)$}]{
   \includegraphics[width=0.38\linewidth]{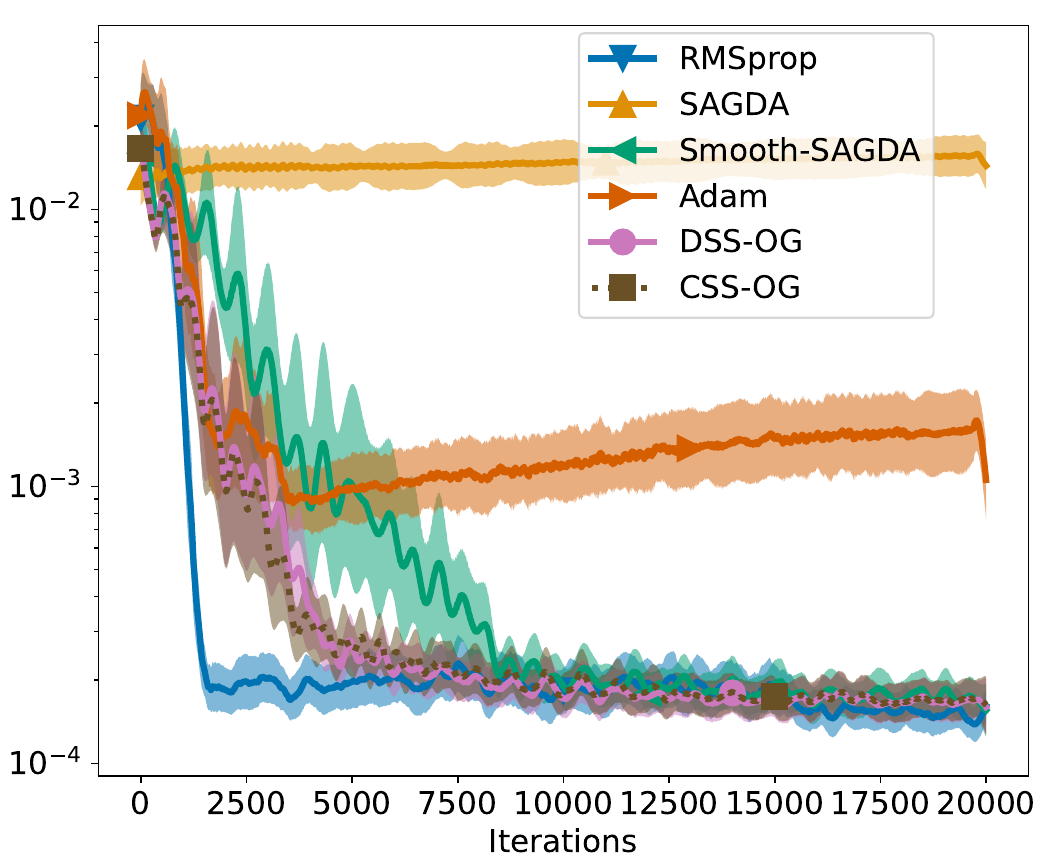}
   \label{fig:example1_variance01_mse}
 }
\caption{Evoluation of gradient norm and mean square error
distance between the true and estimated ones:
In (a) and (b), the true model  is given by $\pi_k =0, \sigma^2_k = 0.001$ for all $k$;
In (c) and (d), the true model is given by $\pi_k =0, \sigma^2_k = 0.1$ for all $k$.
}
\label{fig:sample_figure}
\end{figure*}
\subsection{Wasserstein GAN}
In this example,
we consider the application of \textcolor{black}{an} $\ell_2$-regularized Wasserstein GAN (WGAN)  for learning the mean and variance of a {\em one-dimensional} Gaussian  \cite{yang2022faster}.
The objective is to optimize the following global risk function through the cooperation of agents:
\begin{align} 
\label{bilinear}
&\min_{{x}} \max_{{y}} \  J({x}, {y}) =\sum_{k=1}^{K} p_k J_k({x}, {y}) \\
&J_k({{x}, {y}})  \triangleq
 \mathbb{E}_{\bd{u}_k \sim \mathcal{N}(\pi_k, \sigma^2_k)}[D({y}; \bd{u}_k)]\notag \\
 &\qquad \qquad-\mathbb{E}_{\bd{z}_k \sim \mathcal{N}(0, I)}[D({y};G({x}; \bd{z}_k))]  - \lambda \|y\|_2 \notag
\end{align}
Here,
the dual variable $y$ is $\ell_2$-regularized  by a constant $\lambda$ to ensure \textcolor{black}{the} PL condition.
We consider a setting \textcolor{black}{where}
the generator has a single hidden layer with 5 neurons,
and the discriminator is parameterized by $y \in \mathbb{R}^{2}$  \textcolor{black}{and is of} the form $D(y;\bd{u}) =y_1 \bd{u}_k + y_2 \bd{u}_k^2$.
As for the true samples, $\bd{u}_k$,
\textcolor{black}{they are} drawn from the distribution $\mathcal{N}(\pi_k, \sigma^2_k)$.

We  generate a strongly connected network with 
$K=8$ agents and use  the averaging rule \cite{sayed2014diffusion,sayed2022inference}.
In addition to \textcolor{black}{the} fully distributed setting,
we also implement centralized 
\textbf{SS-OG} (\textbf{CSS-OG}).
We compare 
\textbf{DSS-OG} and \textbf{CSS-OG}
with 
the distributed version of
SAGDA, \textbf{smoothed} {SAGDA} and the adaptive algorithms, such as 
\textbf{RMSprop} and \textbf{Adam} which are also evaluated in this same application \cite{yang2022faster}. 
For these algorithms,
we follow the optimal settings reported 
in \cite{yang2022faster}.
For  \textbf{DSS-OG} and \textbf{CSS-OG},
we tune the step sizes to $\mu_x = \mu_y = 0.1$.
We set the same random seed before running each algorithm to ensure the same generated training set.
\textcolor{black}{The} code is available at the attached link\footnote{\url{https://github.com/MulGarFed/Minimax_DSSOG.git}}.

The simulation results are demonstrated 
\textcolor{black}{in }  Figs.\ref{fig:example1_variance0001_grad}-\ref{fig:example1_variance01_mse}.
\textcolor{black}{We} observe
both   
\textbf{DSS-OG} and \textbf{CSS-OG}
outperform the other algorithms
in terms of gradient norm and mean square error (MSE)
when $\sigma^2_k = 0.001$.
From Fig. \ref{fig:example1_variance01_mse},
the MSE of \textbf{DSS-OG} and \textbf{CSS-OG}
is able to match the results of \textbf{RMSprop} when $\sigma^2_k = 0.1$.
We \textcolor{black}{find} that the adaptive algorithm \textbf{RMSprop} converges more rapidly than constant step size algorithms.
Another interesting observation is that the gradient norm does not necessarily imply the quality of a solution.  For instance, despite \textbf{RMSprop} not having the lowest gradient norm in Fig. \ref{fig:example1_variance0001_grad}, it
exhibits rapid convergence and good performance in terms of MSE, as seen in Fig. \ref{fig:example1_variance0001_mse}.  This discrepancy suggests a promising research direction in identifying more practical metrics for evaluating minimax solutions.

\begin{figure*}[!htbp]
\centering
\vspace{-1.5em}
\subfigure[FID score over training epochs.]{
   \includegraphics[width=0.34\linewidth]{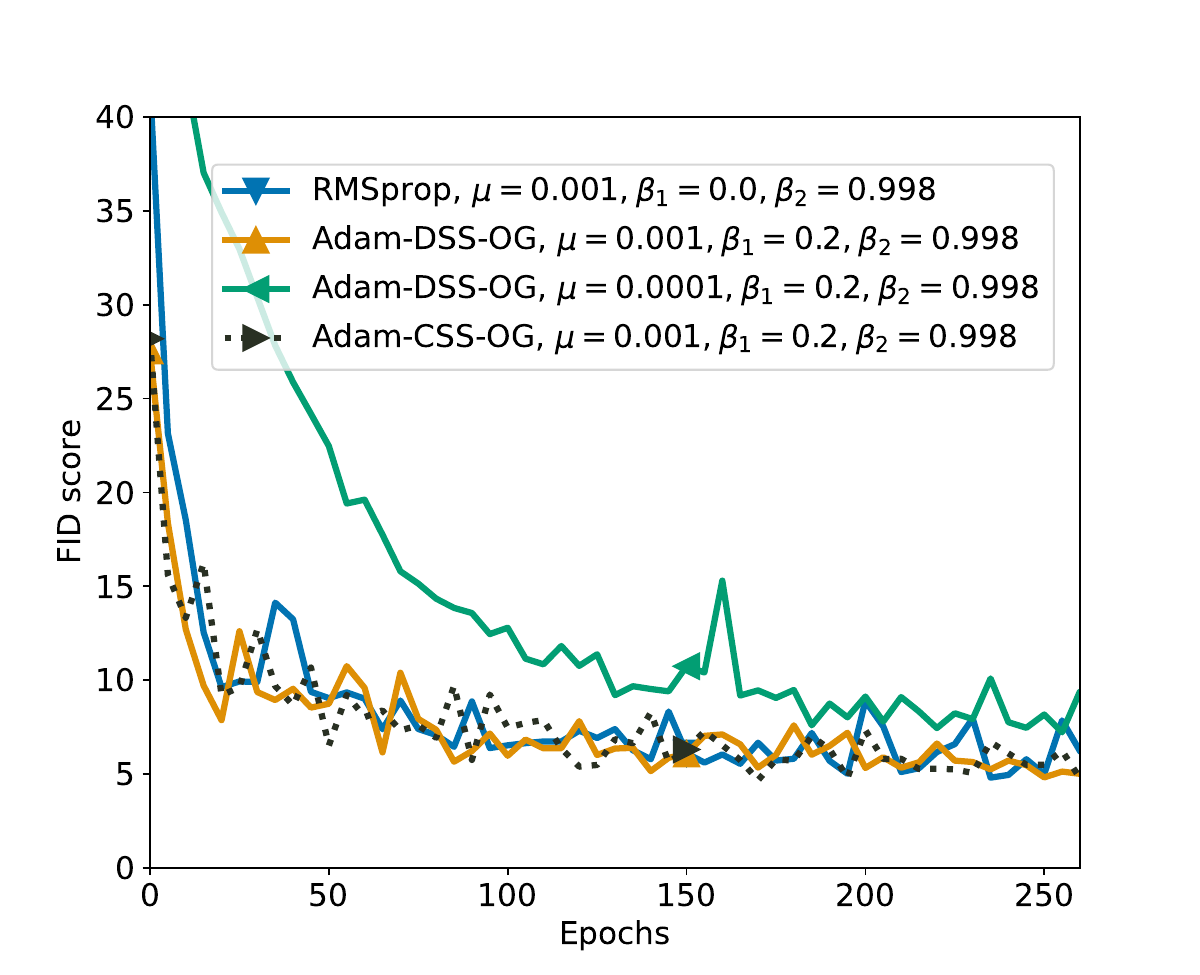}
   \label{fig:fidvalue}
 }
 \hspace{3em}
 \subfigure[Generated images]{
   \includegraphics[width=0.28\linewidth]{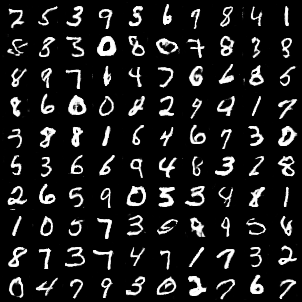}
   \label{fig:epoch_300_agent}
 }
 \caption{Simulation results for DCGANs.
 In (a), the FID score versus different hyperparameters is demonstrated. In (b),
 we show a sample image generated by a single node of \textbf{Adam-DSS-OG} ($\mu = 0.001, \beta_1 = 0.2, \beta_2 = 0.998$).}
  \vspace{-1.5em}
 \end{figure*}
 \vspace{-1em}
\subsection{Deep Convolutional GANs}
In this example, we train a deep convolutional GANs (DCGANs)
using the MNIST dataset.
We consider a similar neural network structure described in 
\cite{radford2015unsupervised}.

We use the same 
network topology from the previous example.
To enhance \textcolor{black}{performance},
we  implement \textbf{DSS-OG}
with a Adam optimizer through replacing its original stochastic gradient term with the \textbf{SS-OG}
gradients from
\eqref{eq7a}--(7b).
For this experiment, we set the momentum factors \textcolor{black}{to} $\beta_1 =0.2, \beta_2 = 0.999$ 
and vary the step sizes.
We plot the results of \textcolor{black}{the} Fr\'echet Inception Distance (FID) score \cite{heusel2017gans}
in Fig. 2(a) The  sample images generated by a single node \textcolor{black}{are} demonstrated in Fig. 2(b).
It is seen that the simulated algorithms can attain good results of FID score 
with the appropriately chosen hyperparameters. Moreover,
the single agent 
 is capable of generating all classes of digits, demonstrating the generative diversity of
a single node in the distributed network.
\section{Conclusion}
In this work, we introduced 
and analyzed \textbf{DSS-OG}, a novel distributed method for stochastic nonconvex-PL minimax optimization. This method improves the analysis of the existing \textcolor{black}{works} by addressing the large batch size issue. We validated the performance of \textbf{DSS-OG} through training experiments of two neural network models: WGAN and DCGAN. 
Our future work is to
develop distributed, communication-efficient, and faster algorithms for 
variational inequality setting under cooperative or competing networks.


%

\appendices
\section{Basic Lemmas}
{\color{black} In this section, we list two basic results that will be used in our analysis.}
\begin{Lemma}[\textbf{Quadratic Growth} \cite{karimi2016linear}]
\label{QGprop}
If $ g(y)$  is $\nu$-PL function, then it also satisfies \textcolor{black}{a}
quadratic growth, i.e.,
\begin{equation}
\begin{aligned}
    g(y) - \min_{y} g(y)
    \ge \frac{\nu}{2}
    \|y^o -y\|^2, \forall y
\end{aligned}
\end{equation}
where $y^o$
is any optimal point taken from the set $\{y^o \mid y^o= {\operatorname{argmin}}_y \  g(y) \}$.
\end{Lemma}
\vspace{-0.5em}
\begin{Lemma}[\textbf{Danskin-type Lemma} \cite{nouiehed2019solving}] \label{Danskin}
Under 
Assumptions \ref{muPL}--\ref{Lipschitz},
if $- J(x, y)$ is $L_f$-smooth and satisfies 
$\nu$-PL in $y$ for any  $x$, 
then:
\begin{itemize}
\item The primal objective  $P(x)$ 
is $L$-smooth \textcolor{black}{with} \scalebox{0.9}{$L = L_f + \frac{L^2_f}{\nu}$}.
\item The gradient of $P(x)$
is equal to the gradient of $J(x, y^o(x))$ with respective to $x$, i.e.,
\begin{equation}
\nabla P(x)
= \nabla_{x} J(x, y^o(x))
\end{equation}
where $y^o(x)$ is a maximum 
point of $J(x, y)$ for a fixed $x$, i.e,
$y^o(x)\in { \operatorname{argmax}}_y \  J(x, y)$.
\end{itemize}
\end{Lemma}
\section{Key Lemmas}
{\color{black}Here, we list the auxiliary lemmas leading to the main convergence result. The proofs of these results are deferred to later appendices.}
\begin{Lemma} \label{networkdeviation}
Under Assumptions \ref{Lipschitz}--\ref{LeftStochastic},
choosing the step sizes 
\begin{align}
\label{stepcondition}
\mu_x \le \mu_y \le \min \{\frac{1- \|\mathcal{J}^\top_{\gamma_1}\|^2}{ 2 (\|\mathcal{J}^\top_{\gamma_1}\|^2 + 160 \rho^2L^2_f)}, 1\}
\end{align}
the iteration averaged network  
deviation is bounded as:
\begin{align}
&\frac{1}{T}\sum_{i=0}^{T-1}
\mathbb{E}\Big[\|{\boldsymbol{\mathcal{X}}}_{i} - 
\boldsymbol{\mathcal{X}}_{c,i}\|^2 
+ \|{\boldsymbol{\mathcal{Y}}}_{i} - 
\boldsymbol{\mathcal{Y}}_{c,i}\|^2
\Big] \notag \\
&\le 
\frac{40\rho^2(4KG^2 + \mu_y\sigma^2)\mu_y}{1-\|\mathcal{J}^\top_{\gamma_1}\|^2}
\end{align}
where $\rho^2  \triangleq \|\mathcal{V}^\top_{R_1}\|^2\|\mathcal{V}_{L_1}\|^2
\|\mathcal{J}^\top_{\gamma_1}\|^2 $ is a constant associated with the network topology, $\sigma^2 = \sum_{k=1}^K \sigma^2_k$
is the sum of local gradient noise variance.
See Appendix \ref{appennetworkdeviation} for proof.
\end{Lemma} 
Choosing suitable step sizes,
we  observe from Lemma \ref{networkdeviation} that the bound of
iteration averaged network deviation  
is dominated by a term on the order of  $\mathcal{O}(\mu_y)$.
\begin{Lemma} \label{iterateincremental}
Under
Assumptions \ref{Lipschitz}--\ref{LeftStochastic}, 
choosing the step sizes \textcolor{black}{as} in Lemma \ref{networkdeviation},
the iteration averaged 
\textcolor{black}{increment} 
is bounded as:
\begin{align}
   \frac{1}{T}& \sum_{i=0}^{T-1}[\mathbb{E}\|\boldsymbol{\mathcal{X}}_{i} -\boldsymbol{\mathcal{X}}_{i-1}\|^2
+ \mathbb{E}\|\boldsymbol{\mathcal{Y}}_{i} -\boldsymbol{\mathcal{Y}}_{i-1}\|^2]
 \notag \\ \le  & 
\frac{240\rho^2 (4KG^2 + \mu_y\sigma^2)\mu_y}{1-\|\mathcal{J}^\top_{\gamma_1}\|^2}
+\frac{3K}{T}\sum_{i=0}^{T-1}
 [\mu_x^2 \mathbb{E}\|{{g}}_{x, c,i-1}\|^2  \notag \\&  + \mu_y^2 \mathbb{E}\|{{g}}_{y, c,i-1}\|^2]
 +60K \sigma^2 \mu^2_y\underset{k}{\max} \ p^2_k
\end{align}
See Appendix \ref{appeniterateincremental} for proof.
\end{Lemma}
We observe that
the
increment is bounded by three terms,
 i.e., the term associated with network deviation on the order of $\mathcal{O}(\mu_y)$, iteration averaged  {\em risk} gradients centroid, and a  bound caused by the gradient noise on the order of $\mathcal{O}(\sigma^2\mu^2_y)$, respectively.
\begin{Lemma} \label{gradientfgcn}
Under
Assumptions \ref{Lipschitz}--\ref{LeftStochastic}, choosing the step sizes \textcolor{black}{as} in Lemma \ref{networkdeviation},
the iteration averaged  gradient disagreement
is bounded as:
\begin{align}
 \frac{1}{T}& \sum_{i=0}^{T-1}
\mathbb{E}\|\nabla_{w} J(\boldsymbol{x}_{c,i}, \boldsymbol{y}_{c,i}) - g_{w,c,i}\|^2 \quad (w = x \text{ or } y)
\end{align}
\begin{align}
 {\le}&   \ \frac{1120L^2_f\rho^2 (4KG^2 + \mu_y\sigma^2)\mu_y}{1-\|\mathcal{J}^\top_{\gamma_1}\|^2} + 240 K \sigma^2 L^2_f \mu^2_y \underset{k}{\max} \ p^2_k \notag \\
 & \ 
+\frac{12KL^2_f}{T}\sum_{i=0}^{T-1}
 [\mu_x^2 \mathbb{E}\|{{g}}_{x, c,i-1}\|^2  + \mu_y^2 \mathbb{E}\|{{g}}_{y, c,i-1}\|^2] \notag
\end{align}
See Appendix \ref{appengradientfgcn} for proof.
\end{Lemma}
Lemma \ref{gradientfgcn}
implies 
how the deviation between the
global risk function at network centroid $(\boldsymbol{x}_{c,i}, \boldsymbol{y}_{c,i})$ and the centroid of local risk gradient is bounded over iterations. We observe that 
this bound 
consists of three terms, namely the iteration averaged risk gradient centroid and two terms on the order of $\mathcal{O}(\mu_y) + \mc{O}(\sigma^2\mu^2_y)$.
In the following lemmas, 
we will 
establish some useful inequalities associated with the Lipschitz continuity of primal objective \eqref{primalobjective} and risk function \eqref{risk}. For simplicity, we use
\begin{align}
d(x, y)
\triangleq P(x)
-  J(x, y)  
\end{align}
to denote the dual optimality gap for simplicity.
\begin{Lemma} \label{smoothprimal}
Under Assumptions \ref{muPL}--\ref{LeftStochastic}, the  primal objective $P(x)$
 satisfies:
\begin{align}
  \mathbb{E}&[P(\boldsymbol{x}_{c, i+1})] -\mathbb{E}[{{P}}(\boldsymbol{x}_{c, i})] \notag
\\
\le  &
 -
\frac{\mu_x}{2}
\mathbb{E}\|\nabla {{P}}(\boldsymbol{x}_{c,i})\|^2
-
\frac{\mu_x}{2}(1- L\mu_x)\mathbb{E}\|{g}_{x,c,i}\|^2
\notag \\&+\frac{2\mu_xL^2_f}{\nu}
\mathbb{E}[{d}_c(i)]  +\mu_x
\mathbb{E}\|\nabla_x J(\boldsymbol{x}_{c,i}, \boldsymbol{y}_{c,i}) - {g}_{x,c,i}\|^2 
\notag \\
&  
+ 5L\mu^2_x\sigma^2\underset{k}{\max} \ p^2_k 
\end{align}
where ${d}_{c}(i) \triangleq {d}(\boldsymbol{x}_{c, i}, \boldsymbol{y}_{c,i}) =P(\boldsymbol{x}_{c, i})
-  J(\boldsymbol{x}_{c, i}, \boldsymbol{y}_{c, i})
$  denotes the  dual optimality gap evaluated at the network centroid $(\boldsymbol{x}_{c, i}, \boldsymbol{y}_{c,i})$.
See Appendix \ref{proofsmoothprimal} for proof.
\end{Lemma} 
\begin{Lemma} \label{smoothJxJy}
Under Assumptions \ref{muPL}--\ref{LeftStochastic}, we have
\begin{align}
\label{Smoothproof1}
     \mathbb{E}& [-J(\boldsymbol{x}_{c, i+1}, \boldsymbol{y}_{c,i+1})] \\
 \le   &
\mathbb{E}[- J(\boldsymbol{x}_{c, i+1}, \boldsymbol{y}_{c, i})]
- \frac{\mu_y}{2}\mathbb{E}\|\nabla_{y} J(\boldsymbol{x}_{c, i+1}, \boldsymbol{y}_{c, i})\|^2 \notag
\\&-
\frac{\mu_y(1-L_f\mu_y)}{2}\mathbb{E}\|{g}_{y,c,i}\|^2
+ \mu_yL^2_f\mathbb{E}\|\boldsymbol{x}_{c, i+1}-
\boldsymbol{x}_{c, i}\|^2 \notag
 \\
&  +\mu_y\mathbb{E}\|\nabla_{y} J(\boldsymbol{x}_{c, i}, \boldsymbol{y}_{c, i})-
{g}_{y,c,i}\|^2+ 5L_f\mu^2_y \sigma^2 \underset{k}{\max} \ p^2_k \notag
\end{align}
and
\begin{align}
\label{Smoothproof2}
  \mathbb{E}&[ -J(\boldsymbol{x}_{c, i+1}, \boldsymbol{y}_{c, i})]
\\
\le &  \mathbb{E}[-J(\boldsymbol{x}_{c, i}, \boldsymbol{y}_{c, i})]
+ 
\frac{\mu_x}{2}\mathbb{E}\|\nabla_{x} J(\boldsymbol{x}_{c, i}, \boldsymbol{y}_{c, i})- {g}_{x,c,i}\|^2 \notag
\\&+ \frac{\mu_x(3+L_f\mu_x)}{2}\mathbb{E}\|{g}_{x,c,i}\|^2
+ 5L_f\mu^2_x \sigma^2 \underset{k}{\max} \ p^2_k \notag
\end{align}
See Appendix \ref{proofsmoothJxJy} for proof.
\end{Lemma}
\begin{Lemma} \label{dualgap}
Under Assumptions \ref{muPL}--\ref{LeftStochastic},
choosing step sizes 
\begin{equation}
    \scalebox{1.0}{$\mu_x \le \min \{\mu_y, \frac{\nu^2\mu_y}{4L^2_f}, \sqrt{\frac{L_f}{L+L_f}} \mu_y\},
\mu_y \le \min \{1, \frac{1}{\nu}, \frac{1}{L_f}\}$}
\end{equation}
the iteration averaged  dual optimality gap 
is bounded as:
\begin{align}\label{mainresultbn}
  \frac{1}{T} &\sum_{i=0}^{T-1} \mathbb{E}[ {d}_{c}(i)] \notag
  \\
   \le &
  \frac{2}{\nu\mu_y T}{d}_{c}(0)
  + 
\frac{1}{T}
\sum_{i=0}^{T-1}\Big\{
  \frac{3\mu_x(1-\nu\mu_y)}{\nu\mu_y}
  \mathbb{E}\|\nabla_{x} J(\boldsymbol{x}_{c, i}, \boldsymbol{y}_{c, i}) \notag
  \end{align}
  \begin{align}&-
{g}_{x,c,i}\|^2 + 
(1-\nu\mu_y)\Big[\frac{2\mu_x}{\nu\mu_y} + \frac{(L+L_f)\mu^2_x}{\nu\mu_y}\Big]\mathbb{E}\|{g}_{x,c,i}\|^2\notag\\
& +\frac{2L_f^2}{\nu }\mathbb{E}\|\boldsymbol{x}_{c, i+1}-\boldsymbol{x}_{c, i}\|^2
+\frac{2}{\nu} \mathbb{E}\| \nabla_{y}J(\boldsymbol{x}_{c, i}, \boldsymbol{y}_{c, i}) - {g}_{y,c,i}\|^2 \Big\} \notag \\
&- \frac{\mu_y}{2T}(1 - L_f\mu_y)
\sum_{i=1}^{T-1} \sum_{j=0}^{i-1}
(1 - \frac{\nu\mu_y}{2})^{i-j}
\mathbb{E}\|{g}_{y,c,j}\|^2 \notag \\
& +\frac{20L_f\mu_y\sigma^2}{\nu } \underset{k}{\max} \  p^2_k 
\end{align}
See Appendix \ref{Apendualgap} for proof.
\end{Lemma}
The bound of dual optimality gap
is quantified  by several terms, including the initialization term  
which decays on the rate of $\mathcal{O}(\frac{1}{\mu_yT})$, the gradient disagreement terms,  the {\em risk} gradients centroid, the increment of network centroid,
and a constant term on the order of $\mc{O}(\sigma^2\mu_y)$.


\section{Proof of Theorem 1}
\label{proofoftheorem1}
 We now prove the best iterate convergence for the primal variable.
Moving the gradient term
$\mathbb{E}\|\nabla P(\boldsymbol{x}_{c,i})\|^2$ to the left hand side in Lemma \ref{smoothprimal}
and averaging the inequality over iterations,
we get
\begin{align}  
 \frac{1}{T} &\sum_{i=0}^{T-1}   \mathbb{E}\|\nabla P(\boldsymbol{x}_{c,i})\|^2 \notag  
\\
 \le  &  \frac{1}{ T}
\sum_{i=0}^{T-1} \Big\{
\frac{2}{\mu_x}\mathbb{E}[P(\boldsymbol{x}_{c, i})-P(\boldsymbol{x}_{c, i+1}) ]
-(1-L\mu_x)\mathbb{E}
\|{g}_{x,c,i}\|^2 \notag \\&+ \frac{4L_f^2}{\nu}
\mathbb{E}[
{d}_{c}(i)]+ 2
\mathbb{E}\|\nabla_{x} J(\boldsymbol{x}_{c,i}, \boldsymbol{y}_{c,i}) - {g}_{x,c,i}\|^2  \Big\} \notag\\&+ 10L\mu_x\sigma^2\underset{k}{\max} \ p^2_k \label{smoothstep2} 
\\
\overset{(a)}{\le}&    \frac{2}{\mu_x T}
[P(\boldsymbol{x}_{c, 0})-P^\star ]
-\frac{1}{T}\sum_{i=0}^{T-1} 
\Big\{ (1-L\mu_x)
\mathbb{E}\|{g}_{x,c,i}\|^2
\notag \\ &- \frac{4L_f^2}{\nu}
\mathbb{E}[{d}_{c}(i)]- 2
\mathbb{E}\|\nabla_{x} J(\boldsymbol{x}_{c,i}, \boldsymbol{y}_{c,i}) - {g}_{x,c,i}\|^2 
\Big\}\notag\\
&+ 10L\mu_x\sigma^2\underset{k}{\max} \ p^2_k
\label{smoothstep22}
\end{align}
where $(a)$ follows from 
Assumption \ref{boundPhi} and
the telescoping results 
of  $\sum_{i=0}^T P(\boldsymbol{x}_{c, i})-P(\boldsymbol{x}_{c, i+1})$
in \eqref{smoothstep2}.
Invoking Lemma \ref{dualgap}
 and
regrouping the terms in \eqref{smoothstep22},
we obtain  
\begin{align} \label{sumPxc}
 \frac{1}{T}& \sum_{i=0}^{T-1}  \mathbb{E} \|\nabla P(\boldsymbol{x}_{c,i})\|^2  
\\
\le    &
\frac{2}{\mu_x T}
[P(\boldsymbol{x}_{c, 0})-P^\star ]-
\frac{1}{T}\sum_{i=0}^{T-1} 
\Bigg\{\Big\{ 1-L\mu_x
-\frac{4(1-\nu\mu_y)L^2_f}{\nu}
\notag\\& \times \Big(\frac{2\mu_x}{\nu\mu_y} + \frac{(L+L_f)\mu^2_x}{\nu\mu_y}\Big)\Big\}
\mathbb{E}\|{g}_{x,c,i}\|^2-\Big(\frac{12\mu_x(1-\nu\mu_y)L^2_f}{\nu^2\mu_y} 
\notag\\ & 
+ 2\Big) \mathbb{E}\|\nabla_{x} J(\boldsymbol{x}_{c, i}, \boldsymbol{y}_{c, i}) -
{g}_{x,c,i}\|^2 -{\frac{8L_f^4}{\nu^2}
\mathbb{E}\|\boldsymbol{x}_{c, i+1}-\boldsymbol{x}_{c, i}\|^2}\notag \\
& 
-{\frac{8L_f^2}{\nu^2}
\mathbb{E}\| \nabla_{y}J(\boldsymbol{x}_{c, i}, \boldsymbol{y}_{c, i}) - {g}_{y,c,i}\|^2}\Bigg\}
+ \frac{8L_f^2}{\nu^2\mu_y T}
{d}_{c}(0)\notag \\
&-\frac{2\mu_y L^2_f}{\nu T}(1 - L_f\mu_y)
\sum_{i=1}^{T-1} \sum_{j=0}^{i-1}
(1-\frac{\nu\mu_y}{2})^{i-j}
\mathbb{E}\|{g}_{y,c,j}\|^2  \notag
\\
& 
 + \frac{80L^3_f\mu_y\sigma^2}{\nu^2} \underset{k}{\max} \  p^2_k +  10L\mu_x\sigma^2\underset{k}{\max} \ p^2_k  
   \notag
\end{align}
\begin{align}
\overset{(a)}{\le}    &
\frac{2}{\mu_x T}
[P(\boldsymbol{x}_{c, 0})-P^\star ]-
\frac{1}{T}\sum_{i=0}^{T-1} 
\Bigg\{\Big\{ 1-L\mu_x- \frac{4(1-\nu\mu_y)L^2_f}{\nu}
\notag\\& \times \Big(\frac{2\mu_x}{\nu\mu_y} + \frac{(L+L_f)\mu^2_x}{\nu\mu_y}\Big) -
\frac{8L_f^4\mu^2_x}{\nu^2}\Big\}\mathbb{E}\|{g}_{x,c,i}\|^2  \notag \\ &
 -4\mathbb{E}\|\nabla_{x} J(\boldsymbol{x}_{c, i}, \boldsymbol{y}_{c, i})-
{g}_{x,c,i}\|^2  - \frac{8L_f^2}{\nu^2}
\notag
\\&{ \times 
\mathbb{E}\| \nabla_{y}J(\boldsymbol{x}_{c, i}, \boldsymbol{y}_{c, i}) - {g}_{y,c,i}\|^2}
\Bigg\}  +
\frac{80 L^4_f\mu^2_x \sigma^2 }{\nu^2} \underset{k}{\max}  \ p^2_k \notag
\\
&+ \frac{8L_f^2}{\nu^2\mu_y T}
{d}_{c}(0) + \frac{80L^3_f\mu_y\sigma^2}{\nu^2} \underset{k}{\max} \  p^2_k +  10L\mu_x\sigma^2\underset{k}{\max} \ p^2_k
  \notag 
\\&
-\frac{2\mu_y L^2_f}{\nu T}(1 - L_f\mu_y)
\sum_{i=1}^{T-1} \sum_{j=0}^{i-1}
(1-\frac{\nu\mu_y}{2})^{i-j}
\mathbb{E}\|{g}_{y,c,j}\|^2   \notag
\end{align}
\textcolor{black}{where} $(a)$
follows from 
several steps:
inserting $\bd{x}_{i+1} = \bd{x}_{i} -
\mu_x \bd{g}_{x,c,i}$, 
using \eqref{gxifi} and choosing step size 
\begin{equation}
    \scalebox{1.0}{$\mu_y \le \frac{1}{\nu}, \frac{12\mu_x(1-\nu\mu_y)L^2_f}{\nu^2\mu_y} \le 2 \Rightarrow
\mu_x \le \frac{\nu^2 \mu_y}{6L^2_f}$}
\end{equation}  
For the sake of simplicity, we introduce the following notations for constants:
\begin{equation}
\scalebox{0.9}{$
\begin{aligned}
\tau_{1} &\triangleq \frac{80L^4_f}{\nu^2},
\tau_{2} \triangleq \frac{80L^3_f}{\nu^2}, \tau_3 \triangleq 
\frac{8L^2_f}{\nu^2}, \tau_4 \triangleq 10L \\
\beta_1 &\triangleq \frac{96KL_f^4}{\nu^2} + 48KL^2_f
\end{aligned}
$}
\end{equation}
Let $P_0 = P(\boldsymbol{x}_{c, 0})-P^\star$ be the initial  primal optimality gap. Invoking 
Lemma \ref{gradientfgcn}
for inequality \eqref{sumPxc}, we get
\begin{align} 
 \frac{1}{T}&\sum_{i=0}^{T-1}   \mathbb{E}\|\nabla P(\boldsymbol{x}_{c,i})\|^2   \\
\le  & 
\frac{2P_0}{\mu_x T} 
+
(\tau_2\mu_y+\tau_4\mu_x+\tau_1\mu^2_x)\sigma^2\underset{k}{\max} \ p^2_k  +  \frac{\tau_3{d}_{c}(0)}{\mu_y T}
\notag \\&
 - \frac{1}{T} \Big\{1-L\mu_x
- \frac{4(1-\nu\mu_y)L^2_f}{\nu}\Big(\frac{2\mu_x}{\nu\mu_y} + \frac{(L+L_f)\mu^2_x}{\nu\mu_y}\Big)  \notag
\\&-
\tau_1\mu^2_x - \beta_1\mu^2_x
\Big\}
\sum_{i=0}^{T-1}\underbrace{\mathbb{E}\|{g}_{x,c,i}\|^2}_{\text{primal gradient}}+\frac{\beta_1 \mu_y^2}{T}  \sum_{i=0}^{T-1}
\mathbb{E}\underbrace{\|{{g}}_{y, c,i-1}\|^2}_{\text{dual gradient}}
   \notag
\\& 
 + 
240 K \sigma^2 L^2_f \mu^2_y \underset{k}{\max} \  p^2_k(\tau_3 + 4 ) - \frac{2\mu_y L^2_f}{\nu T}(1 - L_f\mu_y)\notag
\\& \times
\sum_{i=1}^{T-1} \sum_{j=0}^{i-1}
(1-\frac{\nu\mu_y}{2})^{i-j}
\mathbb{E}\underbrace{\|{g}_{y,c,j}\|^2}_{\text{dual gradient}} + \frac{1120L^2_f\rho^2 \mu_y(\tau_3 + 4)}{1-\|\mathcal{J}^\top_{\gamma_1}\|^2} \notag \\z
& \times   (4KG^2 + \mu_y\sigma^2)+ 
\beta_1\mu^2_x\|g_{x,c,-1}\|^2 
\notag 
\end{align}   
For the terms associated with primal gradient,
we can bound them by choosing a suitable step size:
\begin{align}
&1-L\mu_x
- \frac{4(1-\nu\mu_y)L^2_f}{\nu}\Big(\frac{2\mu_x}{\nu\mu_y} + \frac{(L+L_f)\mu^2_x}{\nu\mu_y}\Big)-
(\tau_1 + \beta_1)
\notag \\ &\times \mu^2_x
\overset{(a)}{\ge}  
1-L\mu_x
- 2\tau_3\frac{\mu_x}{\mu_y}  - \Big(\tau_1+\beta_1 \Big) \mu^2_x \overset{(b)}{\ge} 0
   \label{inequality1}
\end{align}
where $(a)$ we choose $ \mu_x \le 2/(L+L_f)$, (b)
we  choose $\mu_x \le \frac{1}{4L},
\mu_x \le \frac{\mu_y}{4\tau_3}$
 and $
\mu^2_x \le 1/(4(\tau_1+\beta_1))
$.
By doing so, we can eliminate the primal gradient terms since now it has a negative sign.
For  terms associated with dual gradient,
we can bound it by deducing that: 
\begin{align}
& \frac{2\mu_yL^2_f}{\nu T}(1 - L_f\mu_y)
\sum_{i=1}^{T-1}
\sum_{j=0}^{i-1}
(1-\frac{\nu\mu_y}{2})^{i-j}
\mathbb{E}\|{g}_{y,c,j}\|^2 \notag
\\&-
\frac{\beta_1\mu^2_y}{T}\sum_{i=0}^{T-1} \mathbb{E}\|{{g}}_{y, c,i-1}\|^2 \notag \hspace{20em}
\\
&\overset{(a)}{=} 
\frac{4L^2_f}{\nu^2 T}(1 - L_f\mu_y)
(1 - \frac{\nu\mu_y}{2})
\sum_{i=0}^{T-2}
{(1 - (1 - \frac{\nu\mu_y}{2})^{T-1-i})}
\notag\\&  \quad \times \mathbb{E}\|{g}_{y,c,i}\|^2 -
\frac{\beta_1\mu_y^2}{T} \sum_{i=0}^{T-1}
 \mathbb{E}\|{{g}}_{y, c,i-1}\|^2 \notag 
\\
 &\overset{(b)}{\ge}  \
 \frac{\tau_3}{2 T}(1 - L_f\mu_y)
(1 - \frac{\nu\mu_y}{2})
\sum_{i=0}^{T-2}
{(1 - (1 - \frac{\nu\mu_y}{2}))}
\notag \\& \quad \times \mathbb{E}\|{g}_{y,c,i}\|^2 -
\frac{\beta_1 \mu_y^2}{T}\sum_{i=-1}^{T-2}
 \mathbb{E}\|{{g}}_{y,c,i}\|^2 \notag\\
 &\ge  \
\frac{1}{T} \{
 \frac{2L^2_f}{\nu}(1 - L_f\mu_y)
(1 - \frac{\nu\mu_y}{2})\mu_y
- \beta_1 \mu_y^2\}  \sum_{i=-1}^{T-2}
 \mathbb{E}\|{{g}}_{y,c,i}\|^2 \notag  \\& \quad  - 
 \frac{\beta_1\mu_y^2 \|{{g}}_{y,c,-1}\|^2}{T} \quad \overset{(c)}{\ge}   - 
\frac{\beta_1\mu_y^2 \|{{g}}_{y,c,-1}\|^2}{T}
\label{inequality2}
\end{align}
where $(a)$ we have used 
\begin{align} \label{boundeq}
\sum_{i= 1}^{T-1} \sum_{j= 0}^{i-1}
    q^{i-j}a_j
    = 
q\sum_{i= 1}^{T-1} \sum_{j= 0}^{i-1}
    q^{i-1-j}a_j
   & =
    q\sum_{i=0}^{T-2}a_i  
    (\sum_{t=0}^{T-2-i} q^t)  \notag 
     \\&= q (\sum_{i=0}^{T-2}a_i  \frac{1 - q^{T-1-i}}{1-q})  \notag 
\end{align}
$q \in (0, 1)$, 
$(b)$ is due to $1- q^t \ge 1 - q$
when $t\ge1$,
$(c)$ we choose 
$
\mu_y \le \frac{2L^2_f}{\nu\beta_1}
$.
Replacing these gradients with the deduced upper bound, we conclude
\begin{align}
& \frac{1}{T}\sum_{i=0}^{T-1} \mathbb{E}\|\nabla P(\boldsymbol{x}_{c,i})\|^2  \notag \\
&\le  \  \frac{2 P_0}{\mu_x T}
+  \frac{\tau_3{d}_{c}(0)}{\mu_y T}
+  (\tau_2\mu_y+\tau_4\mu_x + \tau_1\mu^2_x)\sigma^2\underset{k}{\max} \ p^2_k   \notag\\
& \quad  +  \frac{1120L^2_f\rho^2 (4KG^2 + \mu_y\sigma^2)\mu_y}{1-\|\mathcal{J}^\top_{\gamma_1}\|^2}(\tau_3+4) +  \frac{\beta_1\mu_y^2 \|{{g}}_{y,c,-1}\|^2}{T}
 \notag\\
& \quad   + 
\frac{\beta_1\mu^2_x\|g_{x,c,-1}\|^2}{T} + 
240 K \sigma^2 L^2_f \mu^2_y \underset{k}{\max} \  p^2_k (\tau_3 + 4 )  
\end{align}
Considering all the step size conditions derived above and  choosing
\begin{equation}
    \scalebox{0.9}{$
    \begin{aligned}
       \mu_x &= \min \{
\frac{\nu^2\mu_y}{6L^2_f}, \frac{1}{2(L+L_f)}, \frac{1}{4L}, \frac{\mu_y}{4\tau_3}, \frac{1}{2\sqrt{\tau_1+\beta_1}}, \mu_y
\} \\ 
\mu_y &= \min \{\frac{1}{\nu}, \frac{2L^2_f}{\nu \beta_1}, \frac{1 -\|\mathcal{J}^\top_{\gamma_1}\|^2}{T^{1/2}},1\}
    \end{aligned}$}
\end{equation} 
we  get the non-asymptotic bound as follows:
\begin{align}
\frac{1}{T}\sum_{i=0}^{T-1} \mathbb{E}\|\nabla P(\boldsymbol{x}_{c,i})\|^2
\le  \mathcal{O}\Big(\frac{1}{{T}^{1/2}}\Big) + \mathcal{O}\Big(\frac{1}{T}\Big)  
+ \mathcal{O}\Big(\frac{1}{T^2}\Big)
\end{align}
It is seen that the convergence rate is dominated by $\mathcal{O}\Big(\frac{1}{{T}^{1/2}}\Big)$, therefore
\textbf{DSS-OG} outputs $\varepsilon$-stationary point after $T_0 = \mathcal{O}(\varepsilon^{-4})$ iteration   and gradient evaluation complexity.

\section{Proof of Theorem 2}
\label{proofoftheorem2}
We further prove that \textbf{DSS-OG} can find 
a solution for the dual problem  
by halting the update of primal variable after  ${\boldsymbol{x}}^\star_{c,T_0}$ is found; this can be achieved by setting $\mu_x = 0$ after iteration $T_0$.
For the convenience of analysis,
 we use  ${\boldsymbol{g}}^\star_{y, c,i}$ 
 and ${{g}}^\star_{y, c,i}$ to denote the 
 stochastic and true gradients
 when  their primal variable is fixed at ${\boldsymbol{x}}^\star_{c,T_0}$.
Repeating the arguments in Lemma \ref{smoothJxJy} at point $x ={\boldsymbol{x}}^\star_{c,T_0} $, we get
\begin{align}
\label{startPL}
\mathbb{E}&[
d({\boldsymbol{x}}^\star_{c,T_0}, \boldsymbol{y}_{c,i+1})]   \\
  \notag  \le  &
  (1 - \nu\mu_y) 
 \mathbb{E} [ d({\boldsymbol{x}}^\star_{c,T_0}, \boldsymbol{y}_{c,i}) ]
    +   \frac{\mu_y}{2}\mathbb{E}
 \|\nabla_{y} J({\boldsymbol{x}}^\star_{c,T_0}, \boldsymbol{y}_{c,i}) 
  \\ &   - {{g}}^\star_{y, c,i}\|^2 -
\frac{\mu_y}{2}(1- L_f\mu_y)\mathbb{E}\|{{g}}^\star_{y, c,i}\|^2  
    + 5L_f\mu^2_y\sigma^2 \underset{k}{\max} \ p^2_k \notag
\end{align}
We now
 study recursion
\eqref{startPL} for sufficiently large iteration $i$. Furthermore, we choose the step size $\mu_y = \frac{2T+1}{\nu(T+1)^2}$.
In the following,
we insert this expression of $\mu_y$ 
into the coefficient associated with $\mathbb{E} [ d({\boldsymbol{x}}^\star_{c,T_0}, \boldsymbol{y}_{c,T}) ]$ in \eqref{startPL},
and keep ``$\mu_y$" notation
in the remaining terms to determine additional step size conditions.
Setting $i=T$ in  \eqref{startPL}, we get 
\begin{align} \label{PLrecursion1}
&\mathbb{E}[d({\boldsymbol{x}}^\star_{c,T_0}, \boldsymbol{y}_{c,T+1})]   \\
    \le  &
  \frac{T^2}{(T+1)^2}  
 \mathbb{E}[d({\boldsymbol{x}}^\star_{c,T_0}, \boldsymbol{y}_{c,T})]
    +
  \frac{\mu_y}{2}\Big[ \mathbb{E}
 \|\nabla_{y} J({\boldsymbol{x}}^\star_{c,T_0}, \boldsymbol{y}_{c,T})\notag\\& -{{g}}^\star_{y, c,T}\|^2 -
(1- L_f\mu_y)\mathbb{E}\|{{g}}^\star_{y, c,T}\|^2\Big] 
    + 5L_f\mu^2_y\sigma^2 \underset{k}{\max} \ p^2_k \notag
\end{align}
Multiplying both sides by
$(T+1)^2$, we have
\begin{align} \label{PLrecursion1add}
&(T+1)^2\mathbb{E}[d({\boldsymbol{x}}^\star_{c,T_0}, \boldsymbol{y}_{c,T+1})]   \\
    \le  & \
  T^2  
 \mathbb{E}[d({\boldsymbol{x}}^\star_{c,T_0}, \boldsymbol{y}_{c,T})]
    +
  \frac{\mu_y(T+1)^2}{2}\Big[ \mathbb{E}
 \|\nabla_{y} J({\boldsymbol{x}}^\star_{c,T_0}, \boldsymbol{y}_{c,T})\notag\\& -{{g}}^\star_{y, c,T}\|^2 -
(1- L_f\mu_y)\mathbb{E}\|{{g}}^\star_{y, c,T}\|^2\Big] 
    + 5(T+1)^2L_f\mu^2_y\sigma^2  \notag \\
  &\times   \underset{k}{\max} \ p^2_k \notag
\end{align}
We further denote 
$\delta_i = i^2 \times 
 \mathbb{E}[  P({\boldsymbol{x}}^\star_{c,T_0})-J({\boldsymbol{x}}^\star_{c,T_0}, \boldsymbol{y}_{c,i}) ]$ 
 and \textcolor{black}{obtain} a recursion regarding 
 $\delta_{i}$.
Iterating the recursion of $\delta_{i}$ from $i=T+1$ to $0$, we obtain
 \begin{align} \label{PLboundstep2}
\delta_{T+1}
\le&   \delta_{0}
 + \frac{\mu_y(T+1)^2}{2}\sum_{i=0}^{T}\Big[\mathbb{E}
 \|\nabla_{y} J({\boldsymbol{x}}^\star_{c,T_0}, \boldsymbol{y}_{c,i})-{{g}}^\star_{y, c,i}\|^2 
 \notag \\& -
(1- L_f\mu_y)\mathbb{E}\|{{g}}^\star_{y, c,i}\|^2\Big]  
    + 5 (T+1)^3 L_f\mu^2_y\sigma^2   
    \underset{k}{\max} \ p^2_k 
\end{align}
Repeating the arguments in Lemma \ref{iterateincremental}, we can bound \textcolor{black}{the} 
 iteration averaged gradient deviation as follows:
\begin{align}
\label{dualgradientdeviation}
 &\frac{1}{T+1}\sum_{i=0}^{T}\mathbb{E}
 \|\nabla_{y} J({\boldsymbol{x}}^\star_{c,T_0}, \boldsymbol{y}_{c,i})-{{g}}^\star_{y, c,i}\|^2  \notag
\\
&\le  \frac{1120 L^2_f\rho^2(4KG^2+\mu_y\sigma^2)\mu_y}{(1-\|\mathcal{J}^\top_{\gamma_1}\|^2)}
 + \frac{12KL^2_f}{\nu (T+1)}\sum_{i=0}^{T}
   \mu_y^2   \notag 
\\& \quad  \times \mathbb{E}\|{{g}}^\star_{y, c,i}\|^2+ 240K \sigma^2 L^2_f \mu^2_y \underset{k}{\max} \ p^2_k 
 \end{align}
 Inserting \eqref{dualgradientdeviation}  into \eqref{PLboundstep2}
 and choosing $
    \frac{12KL^2_f \mu^2_y}{\nu} < 1- L_f \mu_y
    \Rightarrow \mu_y < \sqrt{\frac{\nu}{12KL^2_f}}
$
to eliminate gradient terms  in $\eqref{PLboundstep2}$, we  get ($\delta_0 =0$):
\begin{align}
\label{finaldual}
 \delta_{T+1} &\le\frac{560(T+1)^3L^2_f\rho^2(4KG^2+\mu_y\sigma^2)\mu^2_y}{(1-\|\mathcal{J}^\top_{\gamma_1}\|^2)} + 5 (T+1)^3
   \notag \\
 &\quad   \times  L_f\mu^2_y\sigma^2 \underset{k}{\max} \ p^2_k+ 120(T+1)^3K \sigma^2 L^2_f \mu^3_y \underset{k}{\max} \ p^2_k 
\end{align}
Because $\mu_y =\frac{2T+1}{\nu(T+1)^2} \le \frac{2}{\nu(T+1)} $,
dividing both sides of \eqref{finaldual} by $(T+1)^2$,
we get the following non-asymptotic bound for dual optimality gap:
\begin{align}
\mathbb{E}[d({\boldsymbol{x}}^\star_{c,T_0}, \boldsymbol{y}_{c,T+1})] 
&=
 \mathbb{E}[P({\boldsymbol{x}}^\star_{c,T_0})
    - 
   J({\boldsymbol{x}}^\star_{c,T_0}, \boldsymbol{y}_{c,T+1})] 
    \notag \\
&\le \mathcal{O}\Big((T+1)\mu^2_y\Big)
 + \mathcal{O}\Big((T+1)\mu^3_y\Big)    
    \notag
    \\& \le \mathcal{O}\Big(\frac{1}{T+1}\Big) + \mathcal{O}\Big(\frac{1}{(T+1)^2}\Big)
\end{align}
\vspace{-1em}
 \section{Proof of Theorem \ref{maintheorem3}}
 \label{proofoftheorem3}
According to Theorem \ref{maintheorem}, after running $T_0 = \mathcal{O}(\varepsilon^{-4})$ iterations, we can identify the point ${\boldsymbol{x}}^\star_{c, T_0}$ which satisfies the following condition
\begin{equation} 
\begin{aligned}
    \mathbb{E}\|\nabla P({\boldsymbol{x}}^\star_{c, T_0})\|^2 \le  \frac{\varepsilon^2}{4} = \mathcal{O}(\varepsilon^2) 
\end{aligned}
\end{equation}
According to Lemma \ref{Danskin}, this is equivalent to 
\begin{equation} \label{stationary}
\begin{aligned}
\mathbb{E}\|\nabla_{x} J({\boldsymbol{x}}^\star_{c,T_0},
\boldsymbol{y}^o({\boldsymbol{x}}^\star_{c,T_0}))\|^2  \le \frac{\varepsilon^2}{4}
= \mathcal{O}(\varepsilon^2)
\end{aligned}
\end{equation}
\textcolor{black}{Furthermore},
from Theorem \ref{maintheorem2}, after 
iterations $T_1$, we can  find a point 
${\boldsymbol{y}}^\star_{c, T_0+T_1}$
such that
\begin{align} \label{condition1}  
&\mathbb{E}[
  P({\boldsymbol{x}}^\star_{c,T_0})
    - 
    J({\boldsymbol{x}}^\star_{c,T_0}, {\boldsymbol{y}}^\star_{c, T_0+T_1}) ]   \le \frac{1}{L^2_f} \frac{\nu}{8} \varepsilon^{2} = \mathcal{O}(\varepsilon^{2})
\end{align}
We keep the subscript in  
$({\boldsymbol{x}}^\star_{c,T_0},{\boldsymbol{y}}^\star_{{c,T_0+T_1}})$
to indicate the exact iterations at which this point is obtained.
Under condition \eqref{condition1},
we can verify
 $({\boldsymbol{x}}^\star_{c,T_0}, {\boldsymbol{y}}^\star_{c, T_0+T_1})$ 
satisfies the second stationary condition (29b), i.e., 
\begin{align}
\label{secondcondition}
 \mathbb{E}&\|\nabla_{y} J({\boldsymbol{x}}^\star_{c,T_0},
{\boldsymbol{y}}^\star_{c,T_0+T_1}) \|^2
\notag \\
=&    
\mathbb{E}\|\nabla_{y} J({\boldsymbol{x}}^\star_{c,T_0},
{\boldsymbol{y}}^\star_{c,T_0+T_1})
-
\nabla_{y} J({\boldsymbol{x}}^\star_{c,T_0},
\boldsymbol{y}^o({\boldsymbol{x}}^\star_{c,T_0}))\|^2 \notag\\
 \overset{(a)}{\le} &
L^2_f
\mathbb{E}\|{\boldsymbol{y}}^\star_{c,T_0+T_1} - 
\boldsymbol{y}^o({\boldsymbol{x}}^\star_{c,T_0})
\|^2 \notag\\
\overset{(b)}{\le}  & L^2_f
{\frac{2}{\nu}\mathbb{E}[P({\boldsymbol{x}}^\star_{c,T_0})
    - 
    J({\boldsymbol{x}}^\star_{c,T_0}, {\boldsymbol{y}}^\star_{c,T_0+T_1})]}
\notag\\
 \overset{(c)}{\le}  &
\frac{\varepsilon^2}{4} 
\le  \
\varepsilon^2 =\mathcal{O}(\varepsilon^2)
\end{align}
where $(a)$ follows from the 
$L_f$-smooth property of the local risk,
$(b)$ follows from Lemma \ref{QGprop},
$(c)$ follows from
\eqref{condition}.
On the other hand,
the solution $({\boldsymbol{x}}^\star_{c,T_0}, {\boldsymbol{y}}^\star_{c,T_0+T_1})$ also
satisfies the first stationary condition (29a), i.e., 
\begin{align}
  \mathbb{E}&\|\nabla_{x} J({\boldsymbol{x}}^\star_{c,T_0}, 
 {\boldsymbol{y}}^\star_{c,T_0+T_1})\|^2 \notag\\
\le  & 
 \mathbb{E}\|\nabla_{x} J({\boldsymbol{x}}^\star_{c,T_0}, 
 {\boldsymbol{y}}^\star_{c,T_0+T_1}) - 
 \nabla_{x} J({\boldsymbol{x}}^\star_{c,T_0},
 \boldsymbol{y}^o({\boldsymbol{x}}^\star_{c,T_0})) \notag\\&+
\nabla_{x} J({\boldsymbol{x}}^\star_{c,T_0},
 \boldsymbol{y}^o({\boldsymbol{x}}^\star_{c,T_0})) \|^2 \notag\\
\le &
 2\mathbb{E}\|\nabla_{x} J({\boldsymbol{x}}^\star_{c,T_0},
 \boldsymbol{y}^o({\boldsymbol{x}}^\star_{c,T_0}))\|^2 \notag
 \\&+  2\mathbb{E}\|\nabla_{x} J({\boldsymbol{x}}^\star_{c,T_0}, 
 {\boldsymbol{y}}^\star_{c,T_0+T_1}) - 
 \nabla_{x} J({\boldsymbol{x}}^\star_{c,T_0},
 \boldsymbol{y}^o({\boldsymbol{x}}^\star_{c,T_0}))\|^2 \notag
\end{align}
\begin{align}
 \overset{(a)}{\le}&
  2\mathbb{E}\|\nabla_{x} J({\boldsymbol{x}}^\star_{c,T_0},
 \boldsymbol{y}^o({\boldsymbol{x}}^\star_{c,T_0}))\|^2
 \notag\\&+ 2L^2_f
{\frac{2}{\nu}\mathbb{E}[P({\boldsymbol{x}}^\star_{c,T_0})
    - 
    J({\boldsymbol{x}}^\star_{c,T_0}, {\boldsymbol{y}}^\star_{c,T_0+T_1})]}\notag\\
\le  &  \mathcal{O}(\frac{\varepsilon^2}{2}) + \mathcal{O}(\frac{\varepsilon^2}{2}) 
= \mathcal{O}(\varepsilon^2) \hspace{10em}
\end{align}
where $(a)$
follows from \eqref{stationary} and \eqref{condition}.
Therefore, the overall complexity for
iteration and  gradient evaluation  
is given by 
$T_0 + T_1 = \mathcal{O}(\varepsilon^{-4}) +\mathcal{O}(\varepsilon^{-2})$.

The proof is completed here.
\section{Proof of Lemma \ref{networkdeviation}}           
\label{appennetworkdeviation}
Taking conditional expectations of the 
transient term $\|{\boldsymbol{\mathcal{X}}}_{i}- 
\boldsymbol{\mathcal{X}}_{c,i}\|^2 
+ \|{\boldsymbol{\mathcal{Y}}}_{i} - 
\boldsymbol{\mathcal{Y}}_{c,i}\|^2$
over the distribution of 
random samples $\{\boldsymbol{\xi}^x_{k, i}, \boldsymbol{\xi}^y_{k, i}\}_{k=1}^{K}$, we  get
\begin{align}
 \mathbb{E}& \Big[\|{\boldsymbol{\mathcal{X}}}_{i} - 
\boldsymbol{\mathcal{X}}_{c,i}\|^2
+ \|{\boldsymbol{\mathcal{Y}}}_{i} - 
\boldsymbol{\mathcal{Y}}_{c,i}\|^2  \mid \boldsymbol{\mathcal{F}}_{i-1}\Big] \notag
\\
  = &    \mathbb{E}\Big[\|{\boldsymbol{\mathcal{X}}}_{i} - 
(\mathbbm{1} p^\top \otimes I_{M_1}) \boldsymbol{\mathcal{X}}_{i}\|^2 +
\|{\boldsymbol{\mathcal{Y}}}_{i} - 
(\mathbbm{1} p^\top \otimes I_{M_2}) \boldsymbol{\mathcal{Y}}_{i}\|^2 \notag\\&
 \mid  \boldsymbol{\mathcal{F}}_{i-1} \Big] \notag\\
  = & \mathbb{E}\Big[\|[(V^{-1}\otimes I_{M_1})^\top
 (V\otimes I_{M_1})^\top - (\mathbbm{1} p^\top \otimes I_{M_1})]{\boldsymbol{\mathcal{X}}}_{i} 
\|^2 + \| \notag\\  
& [(V^{-1}\otimes I_{M_2})^\top
 (V\otimes I_{M_2})^\top - (\mathbbm{1} p^\top \otimes I_{M_2})]  {\boldsymbol{\mathcal{Y}}}_{i} 
\|^2 \mid   \boldsymbol{\mathcal{F}}_{i-1}  \Big] \notag \\
\overset{(a)}{=} & 
\mathbb{E}\Big[\|\mathcal{V}_{L_1}\mathcal{V}^\top_{R_1}
{\boldsymbol{\mathcal{X}}}_{i} \|^2 + \|\mathcal{V}_{L_2}\mathcal{V}^\top_{R_2}
{\boldsymbol{\mathcal{Y}}}_{i} \|^2 \mid \boldsymbol{\mathcal{F}}_{i-1} \Big] \notag
\\
\overset{(b)}{\le}&  
\|\mathcal{V}_{L_1}\|^2\mathbb{E} \Big[\|\mathcal{V}^\top_{R_1}
{\boldsymbol{\mathcal{X}}}_{i}\|^2 
+\|\mathcal{V}^\top_{R_2}
{\boldsymbol{\mathcal{Y}}}_{i} \|^2 \mid \boldsymbol{\mathcal{F}}_{i-1} 
\Big]  \label{eq:XYcentroid}
\end{align}
where $(a)$ follows from the Jordan decomposition 
of $A$ and
$(b)$ follows from \textcolor{black}{the} sub-multiplicative
property of 
norms and the fact that $\|\mathcal{V}_{L_1}\|^2 = \|\mathcal{V}_{L_2}\|^2$.
Taking expectation over \eqref{eq:XYcentroid} again
and averaging this inequality over iterations, we get:
\begin{align} \label{startbound}
& \frac{1}{T} \sum_{i=0}^{T-1}  \mathbb{E}[\|{\boldsymbol{\mathcal{X}}}_{i} - 
\boldsymbol{\mathcal{X}}_{c,i}\|^2+\| {\boldsymbol{\mathcal{Y}}}_{i} - 
\boldsymbol{\mathcal{Y}}_{c,i}\|^2 ] \notag \\
&\le   \|\mathcal{V}_{L_1}\|^2 \frac{1}{T} \sum_{i=0}^{T-1} \mathbb{E} [ \|\mathcal{V}^\top_{R_1}
{\boldsymbol{\mathcal{X}}}_{i}\|^2
+\|\mathcal{V}^\top_{R_2}
{\boldsymbol{\mathcal{Y}}}_{i} \|^2
] 
\end{align}
In the following, 
we bound the term $\mathbb{E}[ \|\mathcal{V}^\top_{R_1}
{\boldsymbol{\mathcal{X}}}_{i}\|^2]$
by Eqs. \eqref{VrXn}-\eqref{VrXifinal}, $\mathbb{E}[ \|\mathcal{V}^\top_{R_1}
{\boldsymbol{\mathcal{X}}}_{i}\|^2
+\|\mathcal{V}^\top_{R_2}
{\boldsymbol{\mathcal{Y}}}_{i} \|^2
]$ by \eqref{eq:evarageineq}, 
and deduce the final bound \eqref{startbound}
by Eqs. \eqref{eq:betainequality}-\eqref{finalboundXY}.
To bound $\mathbb{E}[ \|\mathcal{V}^\top_{R_1}
{\boldsymbol{\mathcal{X}}}_{i}\|^2]$, we inspect  $\mathbb{E}[\|\mathcal{V}^\top_{R_1}
{\boldsymbol{\mathcal{X}}}_{i} \|^2 \mid \boldsymbol{\mathcal{F}}_{i-1}]$ on the right hand side (RHS) of \eqref{eq:XYcentroid} by deducing that:
\begin{align} 
 \mathbb{E}& \Big[\|\mathcal{V}^\top_{R_1}
{\boldsymbol{\mathcal{X}}}_{i} \|^2 \mid \boldsymbol{\mathcal{F}}_{i-1}\Big] \notag\\
\overset{(a)}{\le} &
\mathbb{E}\Big[\|\mathcal{J}^\top_{\gamma_1}\mathcal{V}^\top_{R_1}[ \boldsymbol{\mathcal{X}}_{i-1} - \mu_x
    \bm{\mathcal{G}}_{x,i-1}]\|^2 \mid \boldsymbol{\mathcal{F}}_{i-1}\Big] \notag \hspace{10em}
\\
  \overset{(b)}{\le}  &
{\|\mathcal{J}^\top_{\gamma_1}\|^2} \|\mathcal{V}^\top_{R_1}{\boldsymbol{\mathcal{X}}}_{i-1}\|^2 
- 2\mu_x {\|\mathcal{J}^\top_{\gamma_1}\|^2}   \mathbb{E} \Big[
\Big\langle \mathcal{V}^\top_{R_1}{\boldsymbol{\mathcal{X}}}_{i-1}, \notag\\
&  \mathcal{V}^\top_{R_1}{\boldsymbol{\mathcal{G}}}_{x,i-1} \Big\rangle \mid 
\boldsymbol{\mathcal{F}}_{i-1}
\Big]
+ \mu^2_x \|\mathcal{J}^\top_{\gamma_1}\|^2  \mathbb{E}\Big[
\|\mathcal{V}^\top_{R_1} {\boldsymbol{\mathcal{G}}}_{x,i-1}\|^2 \mid \boldsymbol{\mathcal{F}}_{i-1}\Big]
\notag 
\\
 \overset{(c)}{\le}&  {\|\mathcal{J}^\top_{\gamma_1}\|^2} \|\mathcal{V}^\top_{R_1}{\boldsymbol{\mathcal{X}}}_{i-1}\|^2 
- 2\mu_x {\|\mathcal{J}^\top_{\gamma_1}\|^2}   \Big\langle \mathcal{V}^\top_{R_1}{\boldsymbol{\mathcal{X}}}_{i-1}, \mathcal{V}^\top_{R_1}{{\mathcal{G}}}_{x,i-1} \Big\rangle 
\notag \\&   + \mu^2_x\|\mathcal{J}^\top_{\gamma_1}\|^2   \mathbb{E}
[\|\mathcal{V}^\top_{R_1}{\boldsymbol{\mathcal{G}}}_{x,i-1}\|^2 \mid \boldsymbol{\mathcal{F}}_{i-1}] \notag 
\end{align}
\begin{align}
\overset{(d)}{\le} & \ {\|\mathcal{J}^\top_{\gamma_1}\|^2} \|\mathcal{V}^\top_{R_1}{\boldsymbol{\mathcal{X}}}_{i-1}\|^2  + \mu_x \|\mathcal{J}^\top_{\gamma_1}\|^2
\Big[\|\mathcal{V}^\top_{R_1}{\boldsymbol{\mathcal{X}}}_{i-1}\|^2  \notag\\& +\|\mathcal{V}^\top_{R_1}{{\mathcal{G}}}_{x,i-1}\|^2 \Big] + 
\mu^2_x\|\mathcal{J}^\top_{\gamma_1}\|^2   \mathbb{E}\Big[|\mathcal{V}^\top_{R_1}{\boldsymbol{\mathcal{G}}}_{x,i-1}\|^2 \mid \boldsymbol{\mathcal{F}}_{i-1}\Big] \notag\\
\overset{(e)}{\le} & {\|\mathcal{J}^\top_{\gamma_1}\|^2} \|\mathcal{V}^\top_{R_1}{\boldsymbol{\mathcal{X}}}_{i-1}\|^2  + \mu_x \|\mathcal{J}^\top_{\gamma_1}\|^2
\Big[\|\mathcal{V}^\top_{R_1}{\boldsymbol{\mathcal{X}}}_{i-1}\|^2  \notag \\& \quad +\|\mathcal{V}^\top_{R_1}{{\mathcal{G}}}_{x,i-1}\|^2 \Big] + 
\mu^2_x \|\mathcal{J}^\top_{\gamma_1}\|^2  
\|\mathcal{V}^\top_{R_1}{{\mathcal{G}}}_{x,i-1}\|^2  \notag \\
 & \quad + 10 \|\mathcal{J}^\top_{\gamma_1}\|^2 \|\mathcal{V}^\top_{R_1}\|^2 \sigma^2 \mu^2_x \quad (\sigma^2 = \sum_{k=1}^{K} \sigma^2_k)  \label{VrXn}
\end{align}
where $(a)$ follows from \eqref{networkupdate} and
$$\mathcal{V}^\top_{R_1} \mathcal{A}_1^\top
=\mathcal{V}^\top_{R_1}(\mathbbm{1} p^\top\otimes I_{M_1} + \mathcal{V}_{L_1} \mathcal{J}^\top_{\gamma_1} \mathcal{V}^\top_{R_1}) = \mathcal{J}^\top_{\gamma_1} \mathcal{V}^\top_{R_1} $$
\textcolor{black}{Moreover,} $(b)$ follows from the sub-multiplicative property of the norm
and the expansion of the squared norm,
$(c)$ follows from the
conditional expectation over the 
sample $\{\boldsymbol{\xi}^{x}_{k,i} \}_{k=1}^{K}$ and the fact that $\boldsymbol{\mathcal{X}}_{i-1}$
only relies on the  random samples up to iteration $i-1$,
$(d)$ follows from the inequality $- 2\langle a, b\rangle \le \|a\|^2 +\|b\|^2$,
$(e)$ 
follows from Assumption \ref{Expectation}  and the following inequality:
\begin{align}
\label{noisebound}
\ \mathbb{E} & 
\Big[\|\mathcal{V}^\top_{R_1}{\boldsymbol{\mathcal{G}}}_{x,i-1}\|^2 \mid \boldsymbol{\mathcal{F}}_{i-1}\Big]  \notag\\
\le &  
\mathbb{E}
\Big[\| \mathcal{V}^\top_{R_1} [
\mbox{\rm col}\{\boldsymbol{s}^x_{k,i-1}\}_{k=1}^{K}
+ \mathcal{G}_{x,i-1}] \|^2
\mid \boldsymbol{\mathcal{F}}_{i-1}
\Big] \notag\\
\le   &
\mathbb{E}\Big[\|\mathcal{V}^\top_{R_1}\mathcal{G}_{x,i-1}\|^2  \mid \boldsymbol{\mathcal{F}}_{i-1}
\Big] + 2\mathcal{V}^\top_{R_1}
\mathbb{E}\Big[
\langle\mathcal{G}_{x,i-1},  \mbox{\rm col}\{\boldsymbol{s}^x_{k,i-1}\}_{k=1}^{K}\rangle 
 \notag\\&  \mid \boldsymbol{\mathcal{F}}_{i-1}\Big]
+
\mathbb{E}[\|\mathcal{V}^\top_{R_1}\mbox{\rm col}\{\boldsymbol{s}^x_{k,i-1}\}_{k=1}^{K}\|^2   \mid \boldsymbol{\mathcal{F}}_{i-1}
] 
\notag\\
\le  & \ 
\|\mathcal{V}^\top_{R_1}\mathcal{G}_{x,i-1}\|^2
 +
10\|\mathcal{V}^\top_{R_1}\|^2 \sum_{k=1}^{K}
\sigma^2_k 
\end{align}
where  $\boldsymbol{s}^x_{k,i}$ denotes the gradient noise
\begin{align}
    \boldsymbol{s}^x_{k,i} &= 2[\nabla_{x} Q_k(\bd{x}_{k, i}, \bd{y}_{k, i};\boldsymbol{\xi}^{x}_{k,i}) -  \nabla_{x} J_k(\bd{x}_{k, i}, \bd{y}_{k, i})] \\
    &\ +[\nabla_{x} Q_k(\bd{x}_{k, i-1}, \bd{y}_{k, i-1};\boldsymbol{\xi}^{x}_{k,i}) -  \nabla_{x} J_k(\bd{x}_{k, i-1}, \bd{y}_{k, i-1})] \notag
\end{align}
Choosing  $\mu_x <1$, we have $\mu^2_x < \mu_x$ and
\begin{align}
\label{boundstep}
\mathbb{E} &\Big[\|\mathcal{V}^\top_{R_1}
{\boldsymbol{\mathcal{X}}}_{i} \|^2 \mid \boldsymbol{\mathcal{F}}_{i-1}] \notag \\ 
\le&  
(1+\mu_x)\ {\|\mathcal{J}^\top_{\gamma_1}\|^2} \|\mathcal{V}^\top_{R_1}{\boldsymbol{\mathcal{X}}}_{i-1}\|^2   + 
2\mu_x\|\mathcal{J}^\top_{\gamma_1}\|^2\|\mathcal{V}^\top_{R_1}{{\mathcal{G}}}_{x,i-1}\|^2 \notag\\
&+ 10 \|\mathcal{J}^\top_{\gamma_1}\|^2 \|\mathcal{V}^\top_{R_1}\|^2 \sigma^2 \mu^2_x  
\end{align}
For the term
$\|\mathcal{V}^\top_{R_1}{{\mathcal{G}}}_{x,i-1}\|^2$,
we can bound it as follows:
\begin{align} 
&\|\mathcal{V}^\top_{R_1}{{\mathcal{G}}}_{x,i-1}\|^2 \notag \hspace{20em}
\\
\overset{(a)}{=} &   \ \Big\|\mathcal{V}^\top_{R_1}\mbox{\rm col}\Big\{2\nabla_x J_k(\bd{x}_{k,i-1}, \bd{y}_{k,i-1})- 
\nabla_x J_k(\bd{x}_{k,i-2}, \bd{y}_{k,i-2})  
\notag \\& \Big\}_{k=1}^K
    - 
\mathcal{V}^\top_{R_1}\mbox{\rm col}\Big\{2\nabla_x J_k(\bd{x}_{c,i-1}, \bd{y}_{c,i-1})\notag - 
\nabla_x J_k(\bd{x}_{c,i-2}, \notag 
\\
& \bd{y}_{c,i-2}) 
\Big\}_{k=1}^K +   \mathcal{V}^\top_{R_1}\mbox{\rm col}\Big\{2\nabla_x J_k(\bd{x}_{c,i-1}, \notag \bd{y}_{c,i-1})
- 
\nabla_x J_k  \notag\\& (\bd{x}_{c,i-2}, \bd{y}_{c,i-2}) 
\Big\}_{k=1}^K  - \mathcal{V}^\top_{R_1} ({\mathbbm{1} p^\top} \otimes I_{M_1})
\mbox{\rm col}\Big\{2\nabla_x J_k
\notag \notag \\  &(\bd{x}_{c,i-1},\bd{y}_{c,i-1})- 
\nabla_x J_k(\bd{x}_{c,i-2}, \bd{y}_{c,i-2}) \Big\}_{k=1}^K 
\Big\|^2\notag
\\
\overset{(b)}{\le} & 
32\|\mathcal{V}^\top_{R_1}\|^2L^2_f\Big[\|\boldsymbol{\mathcal{X}}_{i-1}
- \boldsymbol{\mathcal{X}}_{c, i-1}\|^2  + \|
\boldsymbol{\mathcal{Y}}_{i-1}
- \boldsymbol{\mathcal{Y}}_{c, i-1}\|^2
\Big]
 \notag 
 \\&  + 8\|\mathcal{V}^\top_{R_1}\|^2L^2_f\Big[
\|\boldsymbol{\mathcal{X}}_{i-2}
- \boldsymbol{\mathcal{X}}_{c, i-2}\|^2+\|\boldsymbol{\mathcal{Y}}_{i-2}
- \boldsymbol{\mathcal{Y}}_{c, i-2}\|^2\Big]
 \notag   
\\
& + 20K\|\mathcal{V}^\top_{R_1}\|^2 G^2 \notag
\end{align}
\begin{align}
\label{disagreement2}
\overset{(c)}{\le}  &
32\|\mathcal{V}^\top_{R_1}\|^2\|\mathcal{V}_{L_1}\|^2L^2_f\Big[\|\mathcal{V}^\top_{R_1}\boldsymbol{\mathcal{X}}_{i-1}\|^2 
+\|\mathcal{V}^\top_{R_2}\boldsymbol{\mathcal{Y}}_{i-1}\|^2 \Big] 
 \notag \\&   + 8\|\mathcal{V}^\top_{R_1}\|^2\|\mathcal{V}_{L_1}\|^2L^2_f\Big[\|\mathcal{V}^\top_{R_1}\boldsymbol{\mathcal{X}}_{i-2}\|^2 
+\|\mathcal{V}^\top_{R_2}\boldsymbol{\mathcal{Y}}_{i-2}\|^2\Big] \notag\\ &
 +
20K\|\mathcal{V}^\top_{R_1}\|^2 G^2 
\end{align}
where $(a)$ follows from $\mathcal{V}^\top_{R_1} ({\mathbbm{1} p^\top} \otimes I_{M_1})=0$, $(b)$ follows from Jensen's inequality, \textcolor{black}{the} $L_f$-smooth property of \textcolor{black}{the} local risk $J_k(x, y)$ and Assumption \ref{boundnetwork},
$(c)$ follows from  \eqref{eq:XYcentroid}.
For convenience, we denote the network constant as $\rho^2 \triangleq \|\mathcal{V}^\top_{R_1}\|^2\|\mathcal{V}_{L_1}\|^2
\|\mathcal{J}^\top_{\gamma_1}\|^2=\|\mathcal{V}^\top_{R_2}\|^2\|\mathcal{V}_{L_2}\|^2
\|\mathcal{J}^\top_{\gamma_2}\|^2$. \textcolor{black}{Combining} the results of
\eqref{boundstep}
and \eqref{disagreement2} and taking expectation again,
we get 
\begin{align} \label{VrXifinal} 
   \mathbb{E}&
\|\mathcal{V}^\top_{R_1}
{\boldsymbol{\mathcal{X}}}_{i} \|^2   \\
\le  &
(1+\mu_x)
{\|\mathcal{J}^\top_{\gamma_1}\|^2}
\mathbb{E}
\|\mathcal{V}^\top_{R_1}{\boldsymbol{\mathcal{X}}}_{i-1}\|^2
+ 64\mu_x{\rho}^2 L^2_f\Big[\mathbb{E}\|\mathcal{V}^\top_{R_1}\boldsymbol{\mathcal{X}}_{i-1}\|^2 
 \notag \\ &+\mathbb{E}\|\mathcal{V}^\top_{R_2}\boldsymbol{\mathcal{Y}}_{i-1}\|^2\Big] 
+16\mu_x{\rho}^2L^2_f\Big[\mathbb{E}\|\mathcal{V}^\top_{R_1}\boldsymbol{\mathcal{X}}_{i-2}\|^2
\notag\\ &+\mathbb{E}\|\mathcal{V}^\top_{R_1}\boldsymbol{\mathcal{Y}}_{i-2}\|^2\Big]
+ 10\|\mathcal{V}^\top_{R_1}\|^2\|\mathcal{J}^\top_{\gamma_1}\|^2 (4KG^2 + \mu_x \sigma^2)\mu_x \notag
\end{align}   
\textcolor{black}{A} similar inequality 
holds for $\mathbb{E}
\|\mathcal{V}^\top_{R_2}
{\boldsymbol{\mathcal{Y}}}_{i} \|^2$.
Combining these results together and choosing $\mu_x \le \mu_y$, we have
\begin{align}\label{eq:evarageineq}
\mathbb{E}&[
\|\mathcal{V}^\top_{R_1}
{\boldsymbol{\mathcal{X}}}_{i} \|^2
+ 
\|\mathcal{V}^\top_{R_2}
{\boldsymbol{\mathcal{Y}}}_{i} \|^2] \notag \\
\le & 
(1+\mu_y)
{\|\mathcal{J}^\top_{\gamma_1}\|^2} 
\mathbb{E}[
\|\mathcal{V}^\top_{R_1}{\boldsymbol{\mathcal{X}}}_{i-1}\|^2  + 
\|\mathcal{V}^\top_{R_2}{\boldsymbol{\mathcal{Y}}}_{i-1}\|^2]+ 128\mu_y\rho^2\notag
 \\&   \times L^2_f\mathbb{E}\Big[\|\mathcal{V}^\top_{R_1}\boldsymbol{\mathcal{X}}_{i-1}\|^2 
+\|\mathcal{V}^\top_{R_2}\boldsymbol{\mathcal{Y}}_{i-1}\|^2\Big]   + 32\mu_y\rho^2L^2_f 
 \notag \\&  \times \mathbb{E}\Big[\|\mathcal{V}^\top_{R_1}\boldsymbol{\mathcal{X}}_{i-2}\|^2
+\|\mathcal{V}^\top_{R_2}\boldsymbol{\mathcal{Y}}_{i-2}\|^2\Big] +20\|\mathcal{V}^\top_{R_1}\|^2\|\mathcal{J}^\top_{\gamma_1}\|^2 \notag \\& \times 
 (4KG^2 + \mu_y\sigma^2)\mu_y 
\end{align}
Averaging \eqref{eq:evarageineq} over iterations, we get
\begin{align} \label{eq:betainequality}
&  \frac{1}{T}\sum_{i=0}^{T-1}\mathbb{E}[
\|\mathcal{V}^\top_{R_1}
{\boldsymbol{\mathcal{X}}}_{i} \|^2
+ 
\|\mathcal{V}^\top_{R_2}
{\boldsymbol{\mathcal{Y}}}_{i} \|^2]   
\\
\le  &
\frac{(1+\mu_y)\|\mathcal{J}^\top_{\gamma_1}\|^2+128\mu_y\rho^2L^2_f}{T}
\sum_{i=0}^{T-1}
\mathbb{E}\Big[
\|\mathcal{V}^\top_{R_1}{\boldsymbol{\mathcal{X}}}_{i-1}\|^2 \notag
\\&+
\|\mathcal{V}^\top_{R_2}{\boldsymbol{\mathcal{Y}}}_{i-1}\|^2\Big]
+\frac{32\mu_y\rho^2L^2_f} {T} \sum_{i=0}^{T-1}
\mathbb{E}\Big[\|\mathcal{V}^\top_{R_1}\boldsymbol{\mathcal{X}}_{i-2}\|^2 \notag 
\\&  + \|\mathcal{V}^\top_{R_2}\boldsymbol{\mathcal{Y}}_{i-2}\|^2\Big]
+ 
20\|\mathcal{V}^\top_{R_1}\|^2\|\mathcal{J}^\top_{\gamma_1}\|^2 (4KG^2 + \mu_y\rho^2)\mu_y \notag 
\end{align}
\begin{align}
\overset{(a)}{\le}  & 
\frac{(1 + \mu_y)\|\mathcal{J}^\top_{\gamma_1}\|^2
+ 160\mu_y\rho^2L^2_f}{T} \sum_{i=0}^{T-1}
\mathbb{E}\Big[\|\mathcal{V}^\top_{R_1}\boldsymbol{\mathcal{X}}_{i}\|^2
\notag \\&+\|\mathcal{V}^\top_{R_2}\boldsymbol{\mathcal{Y}}_{i}\|^2\Big]  +
20\|\mathcal{V}^\top_{R_1}\|^2\|\mathcal{J}^\top_{\gamma_1}\|^2 (4KG^2 + \mu_y\sigma^2)\mu_y   \notag 
\end{align}
where  \textcolor{black}{in} $(a)$ we set the initial iterates as $\bd{x}_{k, -1} = \bd{x}_{k, -2} = \bd{y}_{k, -1} = \bd{x}_{k, -2} =0$ and
 add  
 terms $\mathbb{E}\|\mathcal{V}^\top_{R_1}\boldsymbol{\mathcal{X}}_{T-2}\|^2 + \mathbb{E}\|\mathcal{V}^\top_{R_2}\boldsymbol{\mathcal{Y}}_{T-2}\|^2$ and $\mathbb{E}\|\mathcal{V}^\top_{R_1}\boldsymbol{\mathcal{X}}_{T-1}\|^2 + \mathbb{E}\|\mathcal{V}^\top_{R_2}\boldsymbol{\mathcal{Y}}_{T-1}\|^2$ into
the first inequality  to group 
the summation terms.
Since $\|\mathcal{J}^\top_{\gamma_1}\| <1$,
we  choose the step sizes $\mu_y$ such that 
\begin{equation}
\begin{aligned}
&1-  \Big((1+\mu_y)\|\mathcal{J}^\top_{\gamma_1}\|^2
+ 160\mu_y\rho^2L^2_f \Big) \ge \frac{ 1-\|\mathcal{J}^\top_{\gamma_1}\|^2}{2}\\ &
\Longrightarrow \mu_y \le \frac{1- \|\mathcal{J}^\top_{\gamma_1}\|^2}{ 2 (\|\mathcal{J}^\top_{\gamma_1}\|^2 + 160 \rho^2L^2_f)}
\end{aligned}
\end{equation}
Solving inequality \eqref{eq:betainequality} gives
\begin{equation}\label{eqExEy}
\begin{aligned}
&\frac{1}{T}\sum_{i=0}^{T-1}\mathbb{E}[
\|\mathcal{V}^\top_{R_1}
{\boldsymbol{\mathcal{X}}}_{i} \|^2
+ 
\|\mathcal{V}^\top_{R_2}
{\boldsymbol{\mathcal{Y}}}_{i} \|^2] \\
&\le \frac{40\|\mathcal{V}^\top_{R_1}\|^2\|\mathcal{J}^\top_{\gamma_1}\|^2 (4KG^2 + \mu_y\sigma^2)\mu_y}{1-\|\mathcal{J}^\top_{\gamma_1}\|^2}
\end{aligned}
\end{equation}
Combining \eqref{eqExEy} with
\eqref{startbound},
we get the final bound as follows: 
\begin{equation} \label{finalboundXY}
\begin{aligned}
&  \frac{1}{T}\sum_{i=0}^{T-1}
\mathbb{E}[
\|{\boldsymbol{\mathcal{X}}}_{i} - 
\boldsymbol{\mathcal{X}}_{c,i}\|^2 
+ \|{\boldsymbol{\mathcal{Y}}}_{i} - 
\boldsymbol{\mathcal{Y}}_{c,i}\|^2
] \\
& \le  
\frac{40\rho^2 (4KG^2 + \mu_y\sigma^2)\mu_y}{1-\|\mathcal{J}^\top_{\gamma_1}\|^2}
\end{aligned}
\end{equation}
\section{Proof of Lemma \ref{iterateincremental}}
\label{appeniterateincremental}
Note that ${\boldsymbol{\mathcal{X}}}_{c,i} ={\boldsymbol{\mathcal{X}}}_{c, i-1}
     - \mu_x 
     \operatorname{col}\{{\boldsymbol{g}}_{x, c,i-1}\}_{k = 1}^{K}$, we 
insert this expression
 of ${\boldsymbol{\mathcal{X}}}_{c,i}$ into
$\|\boldsymbol{\mathcal{X}}_i - \boldsymbol{\mathcal{X}}_{i-1}\|^2 $ and deduce that
\begin{equation}
\begin{aligned}
      & \|\boldsymbol{\mathcal{X}}_i - \boldsymbol{\mathcal{X}}_{i-1}\|^2  
    \\
    &=  \|\boldsymbol{\mathcal{X}}_i -{\boldsymbol{\mathcal{X}}}_{c,i} 
    + {\boldsymbol{\mathcal{X}}}_{c,i} -
\boldsymbol{\mathcal{X}}_{i-1}
    \|^2   \\
    &=  \|\boldsymbol{\mathcal{X}}_i -{\boldsymbol{\mathcal{X}}}_{c,i} 
    + {\boldsymbol{\mathcal{X}}}_{c, i-1}
     - \mu_x 
     \operatorname{col}\{{\boldsymbol{g}}_{x, c,i-1}\}_{k = 1}^{K}-
    \boldsymbol{\mathcal{X}}_{i-1}
    \|^2  \\
     & \overset{(a)}{\le} 
3\|\boldsymbol{\mathcal{X}}_i -{\boldsymbol{\mathcal{X}}}_{c,i} \|^2 
    +
3K\mu^2_x\|{\boldsymbol{g}}_{x, c,i-1}\|^2 
    +  3\|\boldsymbol{\mathcal{X}}_{i-1} -\boldsymbol{\mathcal{X}}_{c,i-1} \|^2  \hspace{17em}
\end{aligned}
\end{equation}
where $(a)$ follows from Jensen's inequality. \textcolor{black}{The term}
$\|\boldsymbol{\mathcal{Y}}_i - \boldsymbol{\mathcal{Y}}_{i-1}\|^2$
can be bounded similarly, therefore 
\begin{equation} \label{eq:incremental}
\begin{aligned}
  & \|\boldsymbol{\mathcal{X}}_i - \boldsymbol{\mathcal{X}}_{i-1}\|^2  + \|\boldsymbol{\mathcal{Y}}_i -\boldsymbol{\mathcal{Y}}_{i-1}\|^2   \\
  & \le  
3\Big[\|\boldsymbol{\mathcal{X}}_i -{\boldsymbol{\mathcal{X}}}_
{c, i} \|^2 
 +\|\boldsymbol{\mathcal{Y}}_{i} -{\boldsymbol{\mathcal{Y}}}_{c, i} \|^2 \Big]
 +3\Big[\|\boldsymbol{\mathcal{X}}_{i-1} -{\boldsymbol{\mathcal{X}}}_{c, i-1} \|^2
\\& +\|\boldsymbol{\mathcal{Y}}_{i-1} -{\boldsymbol{\mathcal{Y}}}_{c, i-1} \|^2\Big]   + 3 K\Big[\mu_x^2 \|\boldsymbol{g}_{x, c,i-1}\|^2  + \mu_y^2 \|\boldsymbol{g}_{y, c,i-1}\|^2\Big] 
\end{aligned}
\end{equation}
Taking the conditional expectation of \eqref{eq:incremental}
 over the distribution of 
random samples $\{\boldsymbol{\xi}^x_{k, i}, \boldsymbol{\xi}^y_{k, i}\}_{k=1}^{K}$, we get
\begin{equation}
\label{eq:incremental22}
\begin{aligned}
\mathbb{E}&\Big[\|\boldsymbol{\mathcal{X}}_i - \boldsymbol{\mathcal{X}}_{i-1}\|^2 +\|\boldsymbol{\mathcal{Y}}_i -\boldsymbol{\mathcal{Y}}_{i-1}\|^2 \mid 
\boldsymbol{\mathcal{F}}_{i-1}\Big]  \\
 \le &  \  3\mathbb{E}\Big[\|\boldsymbol{\mathcal{X}}_i -{\boldsymbol{\mathcal{X}}}_
{c, i} \|^2 +\|\boldsymbol{\mathcal{Y}}_{i} -{\boldsymbol{\mathcal{Y}}}_{c, i} \|^2 \mid \boldsymbol{\mathcal{F}}_{i-1}\Big]
 \\&+3 \mathbb{E}\Big[\|\boldsymbol{\mathcal{X}}_{i-1} -{\boldsymbol{\mathcal{X}}}_{c, i-1} \|^2
 +  
 \|\boldsymbol{\mathcal{Y}}_{i-1} -{\boldsymbol{\mathcal{Y}}}_{c, i-1} \|^2 \mid \boldsymbol{\mathcal{F}}_{i-1}
 \Big] 
 \\&+ 3K \mathbb{E}\Big[\mu_x^2\|\boldsymbol{g}_{x, c,i-1}\|^2 + \mu_y^2 \|\boldsymbol{g}_{y, c,i-1}\|^2 \mid \boldsymbol{\mathcal{F}}_{i-1}\Big]  
\end{aligned}
\end{equation}
For the term $\mathbb{E}[\|{\boldsymbol{g}}_{x, c,i-1}\|^2|\boldsymbol{\mathcal{F}}_{i-1}]$,
we can bound it as follows: 
\begin{align} \label{gxifi}
   & \ \mathbb{E}[\|{\boldsymbol{g}}_{x, c,i-1}\|^2\mid \boldsymbol{\mathcal{F}}_{i-1}] \\
& = \ 
\mathbb{E}[\|\sum_{k=1}^K p_k (\boldsymbol{s}^x_{k,i-1} + {g}_{x, k,i-1} )\|^2\mid \boldsymbol{\mathcal{F}}_{i-1}] \notag  \hspace{10em}
\\
&\overset{(a)}{=}  \ 
\mathbb{E}\Big[\|\sum_{k=1}^K p_k \boldsymbol{s}^x_{k,i-1}\|^2
+ \|\sum_{k=1}^K p_k {g}_{x, k,i-1} \|^2\mid \boldsymbol{\mathcal{F}}_{i-1}\Big]  \notag
\\
&\overset{(b)}{\le}  \ 
10 \sum_{k=1}^{K} p^2_k \sigma^2_k
+ \| {g}_{x, c,i-1}\|^2 \le  
10 \sigma^2 \underset{k}{\max} \ p^2_k
+ \| {g}_{x, c,i-1}\|^2  \notag
\end{align}
where $(a)$ follows from  Assumption \ref{Expectation} and the fact
the local true gradient ${g}_{x, k,i-1}$
is independent of the gradient noise
$\boldsymbol{s}^x_{k,i-1}$,
$(b)$ follows from 
Assumption \ref{Expectation}.
A similar argument holds for 
 $\mathbb{E}[\|{\boldsymbol{g}}_{y, c,i-1}\|^2\mid \boldsymbol{\mathcal{F}}_{i-1}]$.
Substituting these results 
 into \eqref{eq:incremental22} and \textcolor{black}{choosing} $\mu_x \le \mu_y$, we obtain 
\begin{align}
\label{eq:incremental1} \mathbb{E}&\Big[\|\boldsymbol{\mathcal{X}}_i - \boldsymbol{\mathcal{X}}_{i-1}\|^2  + \|\boldsymbol{\mathcal{Y}}_i -\boldsymbol{\mathcal{Y}}_{i-1}\|^2 \mid \boldsymbol{\mathcal{F}}_{i-1}\Big] \notag \\
 \le  & 
3\mathbb{E}\Big[\|\boldsymbol{\mathcal{X}}_i -{\boldsymbol{\mathcal{X}}}_
{c, i} \|^2 
 +\|\boldsymbol{\mathcal{Y}}_{i} -{\boldsymbol{\mathcal{Y}}}_{c, i} \|^2 \mid \boldsymbol{\mathcal{F}}_{i-1} \Big] \\
 &+3 \mathbb{E}\Big[\|\boldsymbol{\mathcal{X}}_{i-1} -{\boldsymbol{\mathcal{X}}}_{c, i-1} \|^2
 +\|\boldsymbol{\mathcal{Y}}_{i-1} -{\boldsymbol{\mathcal{Y}}}_{c, i-1} \|^2 \mid \boldsymbol{\mathcal{F}}_{i-1} \Big]  
\notag  \\&  + 3K\Big[\mu^2_x
\| {g}_{x, c,i-1}\|^2
+ \mu^2_y
\| {g}_{y, c,i-1}\|^2\Big]
 + 60K \sigma^2 \underset{k}{\max} \ p^2_k
 \mu^2_y \notag
\end{align}
Taking expectations  over 
\eqref{eq:incremental1} again
and averaging the inequality 
over iterations, we get 
\begin{align}
       \frac{1}{T}&\sum_{i=0}^{T-1}
     \mathbb{E} \Big[\|\boldsymbol{\mathcal{X}}_i - \boldsymbol{\mathcal{X}}_{i-1}\|^2 + \|\boldsymbol{\mathcal{Y}}_i -\boldsymbol{\mathcal{Y}}_{i-1}\|^2 \Big] \notag
   \\
    \overset{(a)}{\le} &  \frac{6}{T} \sum_{i=0}^{T-1}     \mathbb{E}\Big[ \|\boldsymbol{\mathcal{X}}_i -{\boldsymbol{\mathcal{X}}}_
{c, i} \|^2 
 +\|\boldsymbol{\mathcal{Y}}_{i} -{\boldsymbol{\mathcal{Y}}}_{c, i} \|^2 \Big] +   \frac{3K}{T}\sum_{i=0}^{T-1} \notag
 \\& 
 \Big[\mu_x^2 \mathbb{E}\|{{g}}_{x, c,i-1}\|^2   + \mu_y^2 \mathbb{E}\|{{g}}_{y, c,i-1}\|^2\Big]  +60K \sigma^2 \underset{k}{\max} p^2_k
 \mu^2_y \notag
\\
\overset{(b)}{\le} &  \frac{240\rho^2 (4KG^2 + \mu_y\sigma^2)\mu_y}{1-\|\mathcal{J}^\top_{\gamma_1}\|^2}
 +\frac{3K}{T}\sum_{i=0}^{T-1}
 \Big[\mu_x^2 \mathbb{E}\|{{g}}_{x, c,i-1}\|^2 
 \notag\\  
 & 
 + \mu_y^2 \mathbb{E}\|{{g}}_{y, c,i-1}\|^2\Big]  +60K \sigma^2 \underset{k}{\max} \ p^2_k
  \mu^2_y
\end{align}   
where $(a)$ is obtained  by adding the positive terms
$\mathbb{E}\|\boldsymbol{\mathcal{X}}_{T-1} -{\boldsymbol{\mathcal{X}}}_{c, T-1} \|^2
+\mathbb{E}\|\boldsymbol{\mathcal{Y}}_{T-1} -{\boldsymbol{\mathcal{Y}}}_{c, T-1} \|^2$
into the RHS of \eqref{eq:incremental1} and  grouping the summation terms,
$(b)$ follows from Lemma \ref{networkdeviation}.
\section{Proof of Lemma \ref{gradientfgcn}}
\label{appengradientfgcn}
Inserting the expression \textcolor{black}{for} $g_{x,c,i}$ into the following deviation term, we get
\begin{align} 
\mathbb{E}&  \|\nabla_x J(\boldsymbol{x}_{c,i}, \boldsymbol{y}_{c,i}) -{g}_{x,c,i}\|^2  \notag \\
= &  
\mathbb{E}\|\nabla_x J(\boldsymbol{x}_{c,i}, \boldsymbol{y}_{c,i}) - 
\sum_{k=1}^{K} p_k[2
\nabla_x J_k(\boldsymbol{x}_{k,i},\boldsymbol{y}_{k,i})
\notag\\&-
\nabla_x J_k(\boldsymbol{x}_{k,i-1},\boldsymbol{y}_{k,i-1})
]
\|^2  \notag \hspace{10em}
\\
\overset{(a)}{\le}  &
2\mathbb{E}\|\nabla_x J(\boldsymbol{x}_{c,i}, \boldsymbol{y}_{c,i}) - \sum_{k=1}^{K}p_k
\nabla_x J_k(\boldsymbol{x}_{k,i},\boldsymbol{y}_{k,i})\Big\|^2 \notag\\&+
2\mathbb{E}\|\sum_{k=1}^{K}p_k[
\nabla_x J_k(\boldsymbol{x}_{k,i},\boldsymbol{y}_{k,i})-
\nabla_x J_k(\boldsymbol{x}_{k,i-1},\boldsymbol{y}_{k,i-1})]\|^2  \notag
\\
\overset{(b)}{\le} & \ 2\sum_{k=1}^{K}
p_k \mathbb{E}
\|\nabla_x J_k(\boldsymbol{x}_{c,i}, \boldsymbol{y}_{c,i})
- \nabla_x J_k(\boldsymbol{x}_{k,i}, \boldsymbol{y}_{k,i})\|^2 \notag\\&+ 2
\sum_{k=1}^{K} p_k
\mathbb{E}
\|\nabla_x J_k(\boldsymbol{x}_{k,i}, \boldsymbol{y}_{k,i})
- \nabla_x J_k(\boldsymbol{x}_{k,i-1}, \boldsymbol{y}_{k,i-1})\|^2 \notag
\\
 \overset{(c)}{\le}&  {4L^2_f}\sum_{k=1}^{K}p_k
\mathbb{E}\Big[\|\boldsymbol{x}_{c,i} -\boldsymbol{x}_{k, i}\|^2
+ \|\boldsymbol{y}_{c,i} -\boldsymbol{y}_{k, i}\|^2
\Big] \notag\\&+
{4L^2_f} \sum_{k=1}^{K}p_k
\mathbb{E}\Big[\|\boldsymbol{x}_{k,i} -\boldsymbol{x}_{k, i-1}\|^2
+ \|\boldsymbol{y}_{k, i} -\boldsymbol{y}_{k, i-1}\|^2\Big] \notag
\end{align}
\begin{align}
\overset{(d)}{\le}&  
4L^2_f \mathbb{E}\Big[ \|\boldsymbol{\mathcal{X}}_i - \boldsymbol{\mathcal{X}}_{c,i}\|^2
+ \|\boldsymbol{\mathcal{Y}}_i - \boldsymbol{\mathcal{Y}}_{c,i}\|^2
\Big] \notag\\&
+ 4L^2_f
  \mathbb{E}\Big[ \|\boldsymbol{\mathcal{X}}_i - \boldsymbol{\mathcal{X}}_{i-1}\|^2
+ \|\boldsymbol{\mathcal{Y}}_i - \boldsymbol{\mathcal{Y}}_{i-1}\|^2 \Big] \hspace{2em}
\label{nablexfxcyc}
\end{align}
where $(a)$  and $(b)$ \textcolor{black}{follow} from
Jensen's inequality,
$(c)$ follows from \textcolor{black}{the}
$L_f$-smooth property of \textcolor{black}{the} local risk function $J_k(x, y)$, $(d)$ is due to $p_k<1$.
Averaging  \eqref{nablexfxcyc}
over iterations, we get 
\begin{align} \label{PLtight}
 \ \frac{1}{T}&\sum_{i=0}^{T-1}
\mathbb{E}\|\nabla_x J(\boldsymbol{x}_{c,i}, \boldsymbol{y}_{c,i}) - {g}_{x,c,i}\|^2  \\
\le  &
\frac{4L^2_f}{T}
\sum_{i=0}^{T-1}
\Big( \mathbb{E}\|\boldsymbol{\mathcal{X}}_i - \boldsymbol{\mathcal{X}}_{c,i}\|^2
+ \mathbb{E}\|\boldsymbol{\mathcal{Y}}_i - \boldsymbol{\mathcal{Y}}_{c,i}\|^2
\Big)\notag
\\&+ \frac{4L^2_f}{T}
\sum_{i=0}^{T-1}
\mathbb{E}\Big[\|\boldsymbol{\mathcal{X}}_i - \boldsymbol{\mathcal{X}}_{i-1}\|^2+
\|\boldsymbol{\mathcal{Y}}_i - \boldsymbol{\mathcal{Y}}_{i-1}\|^2 \Big] \notag
\\
\overset{(a)}{\le} &
\frac{1120L^2_f\rho^2 (4KG^2 + \mu_y\sigma^2)\mu_y}{1-\|\mathcal{J}^\top_{\gamma_1}\|^2} + 240 K \sigma^2 L^2_f \mu^2_y \underset{k}{\max} \ p^2_k \notag \\
 & 
+ \frac{12KL^2_f}{T}\sum_{i=0}^{T-1}
 \Big[\mu_x^2 \mathbb{E}\|{{g}}_{x, c,i-1}\|^2  + \mu_y^2 \mathbb{E}\|{{g}}_{y, c,i-1}\|^2\Big]  \notag
\end{align}
where $(a)$ follows from Lemmas
\ref{networkdeviation}--\ref{iterateincremental}.
\textcolor{black}{A} similar argument holds for $\frac{1}{T}\sum_{i=0}^{T-1}
\mathbb{E}\|\nabla_y J(\boldsymbol{x}_{c,i}, \boldsymbol{y}_{c,i}) - {g}_{y,c,i}\|^2$.
\vspace{-1em}
\section{Proof of Lemma \ref{smoothprimal}}
\vspace{-0.5em}
\label{proofsmoothprimal}
For convenience, we denote
\begin{equation}{d}_c(i) \triangleq {d}(\boldsymbol{x}_{c, i}, \boldsymbol{y}_{c,i}) ={{P}}(\boldsymbol{x}_{c, i}) 
-J(\boldsymbol{x}_{c, i}, \boldsymbol{y}_{c, i})
\end{equation}
with subscript ``c''
to denote the dual optimality gap evaluated at \textcolor{black}{the} network centroid $(\boldsymbol{x}_{c, i}, \boldsymbol{y}_{c,i})$.
From Lemma \ref{Danskin},
$P(x)$
is $L$-smooth, we have 
\begin{align}\label{startsmooth1}
 & \ {{P}}(\boldsymbol{x}_{c, i+1}) \notag\\
  &  \le  \
  {{P}}(\boldsymbol{x}_{c, i})
  + \langle 
  \nabla {{P}}(\boldsymbol{x}_{c,i}),
  \boldsymbol{x}_{c,i+1}
  -
  \boldsymbol{x}_{c,i}
  \rangle
  + \frac{L}{2}\|\boldsymbol{x}_{c,i+1}-  \boldsymbol{x}_{c,i}\|^2  \notag\\
&\le  \ 
{{P}}(\boldsymbol{x}_{c, i}) -
\mu_x \langle 
  \nabla {{P}}(\boldsymbol{x}_{c,i}),
  \boldsymbol{g}_{x,c,i}
  \rangle
  + 
  \frac{L\mu^2_x}{2}\|\boldsymbol{g}_{x,c,i}\|^2
\end{align}
Taking conditional expectation of \eqref{startsmooth1}
over the distribution of random sample 
$\{\boldsymbol{\xi}^x_{k,i+1}\}_{k=1}^{K}$,
we get 
\begin{align}
\label{startsmooth11}
      \ \mathbb{E}&[ {{P}}(\boldsymbol{x}_{c, i+1}) \mid \boldsymbol{\mathcal{F}}_{i}]  \notag\\
\overset{(a)}{\le} & 
{{P}}(\boldsymbol{x}_{c, i}) -
\mu_x \langle 
  \nabla {{P}}(\boldsymbol{x}_{c,i}),
  {g}_{x,c,i}
  \rangle
  + 
  \frac{L\mu^2_x}{2}\mathbb{E}[\|\boldsymbol{g}_{x,c,i}\|^2 \mid \boldsymbol{\mathcal{F}}_{i}] \notag\\
\le  \ & 
{{P}}(\boldsymbol{x}_{c, i}) 
- \frac{\mu_x}{2}  
\|\nabla {{P}}(\boldsymbol{x}_{c, i})\|^2
- \frac{\mu_x}{2}
\|{g}_{x,c,i}\|^2
\notag\\&+ \frac{\mu_x}{2}
\|\nabla{{P}}(\boldsymbol{x}_{c,i})-  {g}_{x,c,i}\|^2
 + 
  \frac{L\mu^2_x}{2}\mathbb{E}[\|\boldsymbol{g}_{x,c,i}\|^2 \mid \boldsymbol{\mathcal{F}}_{i}]  \notag\\
\overset{(b)}{\le} &  {{P}}(\boldsymbol{x}_{c, i}) 
- \frac{\mu_x}{2}
\| \nabla {{P}}(\boldsymbol{x}_{c, i})\|^2
- \frac{\mu_x}{2}
\|{g}_{x,c,i}\|^2 +\frac{\mu_x}{2}
\|\nabla {{P}}(\boldsymbol{x}_{c,i}) \notag\\&  -  {g}_{x,c,i}\|^2
 + 
  \frac{L\mu^2_x}{2}
  (10 \sigma^2 \underset{k}{\max} \ p^2_k + \|{g}_{x,c,i}\|^2)
\end{align}
where $(a)$ follows from the fact that
$\boldsymbol{x}_{c, i}$ does not rely on the  random samples 
at iteration $i+1$,
 $(b)$ follows from \eqref{gxifi}.
Inserting $\nabla_x J(\boldsymbol{x}_{c,i}, \boldsymbol{y}_{c,i})$
into \eqref{startsmooth11}
and applying 
Jensen's  inequality,
we get 
\begin{align}
\label{startsmooth11inter}
  \mathbb{E}&[{{P}}(\boldsymbol{x}_{c, i+1}) \mid \boldsymbol{\mathcal{F}}_{i}] \notag \\
\le  &  
{{P}}(\boldsymbol{x}_{c, i}) -
\frac{\mu_x}{2}
\|\nabla {{P}}(\boldsymbol{x}_{c,i})\|^2
-
\frac{\mu_x}{2}(1- L\mu_x)\|{g}_{x,c,i}\|^2
\notag \\&+\mu_x
\|\nabla {{P}}(\boldsymbol{x}_{c,i})-
\nabla_x J(\boldsymbol{x}_{c,i}, \boldsymbol{y}_{c,i})\|^2  + 5L\mu^2_x\sigma^2 \underset{k}{\max} \ p^2_k \notag \\
& +\mu_x
\|\nabla_x J(\boldsymbol{x}_{c,i}, \boldsymbol{y}_{c,i}) - {g}_{x,c,i}\|^2
\end{align}
Taking expectations again over $\eqref{startsmooth11inter}$, we get 
\vspace{1em}
\begin{align}\label{startsmooth2}
 \ \mathbb{E}&[{{P}}(\boldsymbol{x}_{c, i+1})]  \notag\\
 \le   &
\mathbb{E}[{{P}}(\boldsymbol{x}_{c, i})] -
\frac{\mu_x}{2}
\mathbb{E}\|\nabla {{P}}(\boldsymbol{x}_{c,i})\|^2
-
\frac{\mu_x}{2}(1- L\mu_x)\mathbb{E}\|{g}_{x,c,i}\|^2 \notag
\\&
+\mu_x
\mathbb{E}\|\nabla {{P}}(\boldsymbol{x}_{c,i})-
\nabla_x J(\boldsymbol{x}_{c,i}, \boldsymbol{y}_{c,i})\|^2 \notag \\
& +\mu_x
\mathbb{E}\|\nabla_x J(\boldsymbol{x}_{c,i}, \boldsymbol{y}_{c,i}) - {g}_{x,c,i}\|^2 +  5L\mu^2_x\sigma^2 \underset{k}{\max} \ p^2_k\notag \\
\overset{(a)}{\le} &
\mathbb{E}[{{P}}(\boldsymbol{x}_{c, i})] -
\frac{\mu_x}{2}
\mathbb{E}\|\nabla {{P}}(\boldsymbol{x}_{c,i})\|^2
-
\frac{\mu_x}{2}(1- L\mu_x)\mathbb{E}\|{g}_{x,c,i}\|^2
\notag\\&+\mu_xL^2_f
\mathbb{E}\|
\boldsymbol{y}^{o}(\boldsymbol{x}_{c,i})
-\boldsymbol{y}_{c,i}\|^2 +\mu_x
\mathbb{E}\|\nabla_x J(\boldsymbol{x}_{c,i}, \boldsymbol{y}_{c,i})  
\notag\\&- {g}_{x,c,i}\|^2+ 5L\mu^2_x\sigma^2 \underset{k}{\max} \ p^2_k
\notag\\
\overset{(b)}{\le} &
\mathbb{E}[{{P}}(\boldsymbol{x}_{c, i})] -
\frac{\mu_x}{2}
\mathbb{E}\|\nabla {{P}}(\boldsymbol{x}_{c,i})\|^2
-
\frac{\mu_x}{2}(1- L\mu_x)\mathbb{E}\|{g}_{x,c,i}\|^2
\notag\\&+\frac{2\mu_xL^2_f}{\nu}
\mathbb{E}[{{P}}(\boldsymbol{x}_{c,i})
- J(\boldsymbol{x}_{c,i}, \boldsymbol{y}_{c,i})]  \\
& +\mu_x
\mathbb{E}\|\nabla_x J(\boldsymbol{x}_{c,i}, \boldsymbol{y}_{c,i}) - {g}_{x,c,i}\|^2 + 5L\mu^2_x\sigma^2 \underset{k}{\max} \ p^2_k \notag
\end{align}
\textcolor{black}{Here,} $(a)$ is due to \textcolor{black}{the} $L_f$-smoothness property of \textcolor{black}{the} risk function,
$(b)$ follows from Lemma 1.
\section{Proof of Lemma \ref{smoothJxJy}}
\label{proofsmoothJxJy}
Because $-J(x, y)$
is $L_f$-smooth,  fixing $-J(x, y)$ at $\boldsymbol{x}_{c, i+1}$,
we have
\begin{align} \label{eqfsmooth11}
& -J (\boldsymbol{x}_{c, i+1}, \boldsymbol{y}_{c,i+1})
\notag\\
 \le & - J(\boldsymbol{x}_{c, i+1}, \boldsymbol{y}_{c, i}) -
\langle \nabla_{y} J(\boldsymbol{x}_{c, i+1}, \boldsymbol{y}_{c, i}) , \boldsymbol{y}_{c, i+1} - \boldsymbol{y}_{c, i} \rangle \notag
\\&+ \frac{L_f}{2}\|\boldsymbol{y}_{c, i+1} - \boldsymbol{y}_{c, i}\|^2 \notag
\\
\le  & - J(\boldsymbol{x}_{c, i+1}, \boldsymbol{y}_{c, i}) -
\mu_y \langle \nabla_{y} J(\boldsymbol{x}_{c, i+1}, \boldsymbol{y}_{c, i}) , \boldsymbol{g}_{y,c,i} \rangle
\notag\\&+
\frac{L_f\mu^2_y}{2} \|\boldsymbol{g}_{y,c,i} \|^2   \hspace{10em}
\end{align}
Taking  expectation of \eqref{eqfsmooth11} over the distribution of  
$\{\boldsymbol{\xi}^y_{k,i+1}\}_{k=1}^K$
conditioned on 
$\{\boldsymbol{\xi}^x_{k,i+1}\}_{k=1}^K$
and $\boldsymbol{\mathcal{F}}_{i}$, we get 
\begin{align}
\label{eqfsmoothkey}
 \mathbb{E}&[-J(\boldsymbol{x}_{c, i+1}, \boldsymbol{y}_{c,i+1}) \mid 
\boldsymbol{\mathcal{F}}_{i}, \{\boldsymbol{\xi}^x_{k,i+1}\}_{k=1}^K] \notag
 \\
\le&  \ - J(\boldsymbol{x}_{c, i+1}, \boldsymbol{y}_{c, i}) -
\mu_y \langle \nabla_{y} J(\boldsymbol{x}_{c, i+1}, \boldsymbol{y}_{c, i}) , {g}_{y,c,i} \rangle \notag
\\&+
\frac{L_f\mu^2_y}{2} \mathbb{E}\Big[\|\boldsymbol{g}_{y,c,i} \|^2 \mid \boldsymbol{\mathcal{F}}_{i}, \{\boldsymbol{\xi}^x_{k,i+1}\}_{k=1}^K\Big] \notag
\\
\overset{(a)}{\le} &
- J(\boldsymbol{x}_{c, i+1}, \boldsymbol{y}_{c, i})
- \frac{\mu_y}{2}\|\nabla_{y} J(\boldsymbol{x}_{c, i+1}, \boldsymbol{y}_{c, i})\|^2 
\notag\\&+ \frac{\mu_y}{2}\|\nabla_{y} J(\boldsymbol{x}_{c, i+1}, \boldsymbol{y}_{c, i})-
{g}_{y,c,i}\|^2 + \frac{L_f\mu^2_y}{2}\|{g}_{y,c,i} \|^2 \notag\\
& + 5L_f\mu^2_y \sigma^2 \underset{k}{\max} \ p^2_k -
\frac{\mu_y}{2}\|{g}_{y,c,i}\|^2 \notag
\end{align}
\begin{align}
\overset{(b)}{\le}&
- J(\boldsymbol{x}_{c, i+1}, \boldsymbol{y}_{c, i})
- \frac{\mu_y}{2}\|\nabla_{y} J(\boldsymbol{x}_{c, i+1}, \boldsymbol{y}_{c, i})\|^2 
\notag\\&+ \mu_y\|\nabla_{y} J(\boldsymbol{x}_{c, i+1}, \boldsymbol{y}_{c, i})-
\nabla_{y} J(\boldsymbol{x}_{c, i}, \boldsymbol{y}_{c, i})\|^2  
 \notag\\
&+\mu_y\|\nabla_{y} J(\boldsymbol{x}_{c, i}, \boldsymbol{y}_{c, i})-
{g}_{y,c,i}\|^2
+\frac{L_f\mu^2_y}{2}\|{g}_{y,c,i} \|^2
\notag\\&+ 5L_f\mu^2_y \sigma^2 \underset{k}{\max} \ p^2_k -
\frac{\mu_y}{2}\|{g}_{y,c,i}\|^2 
\notag
\\
\overset{(c)}{\le} &
- J(\boldsymbol{x}_{c, i+1}, \boldsymbol{y}_{c, i})
- \frac{\mu_y}{2}\|\nabla_{y} J(\boldsymbol{x}_{c, i+1}, \boldsymbol{y}_{c, i})\|^2 \\&-
\frac{\mu_y}{2}\|{g}_{y,c,i}\|^2
+ \mu_yL^2_f\|\boldsymbol{x}_{c, i+1}-\boldsymbol{x}_{c, i}\|^2  +\frac{L_f\mu^2_y}{2}\|{g}_{y,c,i} \|^2
 \notag\\
& +\mu_y\|\nabla_{y} J(\boldsymbol{x}_{c, i}, \boldsymbol{y}_{c, i})-
{g}_{y,c,i}\|^2
+ 5L_f\mu^2_y \sigma^2 \underset{k}{\max} \ p^2_k \notag
\end{align}
where $(a)$ follows from 
\eqref{gxifi},
$(b)$  we insert  the true gradient information $\nabla_{y} J(\boldsymbol{x}_{c, i}, \boldsymbol{y}_{c, i})$ and apply Jensen's inequality,
$(c)$ is due to \textcolor{black}{the} $L_f$-smooth property of \textcolor{black}{the} risk function.
Taking expectation again over 
\eqref{eqfsmoothkey}, we obtain 
the result \eqref{Smoothproof1}.
The proof of \eqref{Smoothproof2} is similar to the above arguments.
\vspace{-1em}
\section{Proof of Lemma \ref{dualgap}}
\label{Apendualgap}

 Adding $\mathbb{E}[P(\boldsymbol{x}_{c, i+1})]$ 
 on both sides of \eqref{Smoothproof1}:
\begin{align} \label{eq:recursionforbn} 
 \ \mathbb{E}&[P(\boldsymbol{x}_{c, i+1}) 
-J(\boldsymbol{x}_{c, i+1}, \boldsymbol{y}_{c, i+1})] \notag
\\
\le&  \ \mathbb{E}[P(\boldsymbol{x}_{c, i+1})- J(\boldsymbol{x}_{c, i+1}, \boldsymbol{y}_{c, i})]
- \frac{\mu_y}{2}\mathbb{E}\|\nabla_{y} J(\boldsymbol{x}_{c, i+1}, \boldsymbol{y}_{c, i})\|^2 
\notag \\&-
\frac{\mu_y}{2}(1-L_f\mu_y)\mathbb{E}\|{g}_{y,c,i}\|^2
+ 5L_f\mu^2_y \sigma^2 \underset{k}{\max} \ p^2_k\notag \\
& +  \mu_yL^2_f\mathbb{E}\|\boldsymbol{x}_{c, i+1}-
\boldsymbol{x}_{c, i}\|^2+\mu_y\mathbb{E}\|\nabla_{y} J(\boldsymbol{x}_{c, i}, \boldsymbol{y}_{c, i})-
{g}_{y,c,i}\|^2 \notag 
\\
\overset{(a)}{\le}&   \
(1 - \nu\mu_y) \mathbb{E}[P(\boldsymbol{x}_{c, i+1}) 
-J(\boldsymbol{x}_{c, i+1}, \boldsymbol{y}_{c, i})]
\notag\\ 
&  +\mu_y\mathbb{E}\|\nabla_{y} J(\boldsymbol{x}_{c, i}, \boldsymbol{y}_{c, i})-
{g}_{y,c,i}\|^2+ 5L_f\mu^2_y \sigma^2 \underset{k}{\max} \ p^2_k \notag
\\
&- \frac{\mu_y}{2}(1 - L_f\mu_y)
\mathbb{E}\|{g}_{y,c,i}\|^2 +  \mu_yL^2_f\mathbb{E}\|\boldsymbol{x}_{c, i+1}-
\boldsymbol{x}_{c, i}\|^2\notag
\\ 
\overset{(b)}{\le} &
(1 - \nu\mu_y) \mathbb{E}[
P(\boldsymbol{x}_{c, i})
- J(\boldsymbol{x}_{c, i}, \boldsymbol{y}_{c, i})
+J(\boldsymbol{x}_{c, i}, \boldsymbol{y}_{c, i})
\\&-J(\boldsymbol{x}_{c, i+1}, \boldsymbol{y}_{c, i})
+
P(\boldsymbol{x}_{c, i+1}) 
- P(\boldsymbol{x}_{c, i}) ]
\notag\\& - \frac{\mu_y}{2}(1 - L_f\mu_y)
\mathbb{E}\|{g}_{y,c,i}\|^2
+ {\mu_y}L_f^2\mathbb{E}\|\boldsymbol{x}_{c, i+1}-\boldsymbol{x}_{c, i}\|^2 \notag
\\&+ \mu_y\mathbb{E}\| \nabla_{y}J(\boldsymbol{x}_{c, i}, \boldsymbol{y}_{c, i}) - {g}_{y,c,i}
\|^2  + 5L_f\mu^2_y \sigma^2 \underset{k}{\max} \ p^2_k  \notag
\end{align}   
where $(a)$ follows from Assumption \ref{muPL},
$(b)$ is obtained by adding and subtracting the terms 
$\mathbb{E}[P(\boldsymbol{x}_{c, i})]$ and 
$\mathbb{E}[J(\boldsymbol{x}_{c,i}, \boldsymbol{y}_{c,i})]$ in $(a)$.
By the definition of ${d}_{c}(i)$, we can rewrite above inequality as: 
\begin{align}\label{eq:recursionforbn2}
 \mathbb{E}& [ {d}_{c}(i+1) ]\notag\\
  \le & 
  (1 - \nu\mu_y)
\Big\{ \mathbb{E}[{d}_{c}(i)]
+ \mathbb{E}[J(\boldsymbol{x}_{c, i}, \boldsymbol{y}_{c, i})
-J(\boldsymbol{x}_{c, i+1}, \boldsymbol{y}_{c, i})]
\notag \\
&+
\mathbb{E}[P(\boldsymbol{x}_{c, i+1}) 
- P(\boldsymbol{x}_{c, i}) ] \Big\}+ 5L_f\mu^2_y \sigma^2 \underset{k}{\max} \ p^2_k \notag
\\&  - \frac{\mu_y}{2}(1 - L_f\mu_y) 
\mathbb{E}\|{g}_{y,c,i}\|^2+ {\mu_y}L_f^2\mathbb{E}\|\boldsymbol{x}_{c, i+1}-\boldsymbol{x}_{c, i}\|^2
\notag\\&+ \mu_y\mathbb{E}\| \nabla_{y}J(\boldsymbol{x}_{c, i}, \boldsymbol{y}_{c, i}) - {g}_{y,c,i}
\|^2  
\end{align}
By Lemma \ref{smoothJxJy},
we can bound $\mathbb{E}[J(\boldsymbol{x}_{c, i}, \boldsymbol{y}_{c, i})
-J(\boldsymbol{x}_{c, i+1}, \boldsymbol{y}_{c, i}) ]$
as:
\begin{align} \label{eqfsmooth}
\mathbb{E}&[J(\boldsymbol{x}_{c, i}, \boldsymbol{y}_{c, i})
-J(\boldsymbol{x}_{c, i+1}, \boldsymbol{y}_{c, i}) ] \notag \\
\le &
\frac{\mu_x}{2}\mathbb{E}\|\nabla_{x} J(\boldsymbol{x}_{c, i}, \boldsymbol{y}_{c, i})- {g}_{x,c,i}\|^2
+ \frac{\mu_x(3+L_f\mu_x)}{2}\mathbb{E}\|{g}_{x,c,i}\|^2 \notag
\\&  + 5L_f\mu^2_x \sigma^2 \underset{k}{\max} \ p^2_k
\end{align}   
By Lemma \ref{smoothprimal},
we can bound $\mathbb{E}[{{P}}(\boldsymbol{x}_{c, i+1})-{{P}}(\boldsymbol{x}_{c, i})]$ as
\begin{align} \label{Phixsmooth}
\mathbb{E}&[{{P}}(\boldsymbol{x}_{c, i+1})-{{P}}(\boldsymbol{x}_{c, i})] \notag
\\  \le  &
\frac{2\mu_xL^2_f}{\nu}{d}_{c}(i)-
\frac{\mu_x}{2}(1- L\mu_x)\mathbb{E}\|{g}_{x,c,i}\|^2
\\&+\mu_x
\mathbb{E}\|\nabla_x J(\boldsymbol{x}_{c,i}, \boldsymbol{y}_{c,i}) - {g}_{x,c,i}\|^2  + 5L \mu^2_x\sigma^2 \underset{k}{\max} \  p^2_k \notag
\end{align}
Combining the results of 
\eqref{eqfsmooth}
and \eqref{Phixsmooth} into 
\eqref{eq:recursionforbn2}, 
we get 
\begin{align}\label{eq:recursionforbn3}
\mathbb{E}&[{d}_{c}(i+1)] \\
\le& (1 - \nu\mu_y)(1 + \frac{2\mu_xL_f^2}{\nu})\mathbb{E}[{d}_{c}(i)]
+
\frac{3\mu_x(1-\nu\mu_y)}{2}
\notag\\
&\times \mathbb{E}\|\nabla_xJ(\boldsymbol{x}_{c, i}, \boldsymbol{y}_{c, i}) -
{g}_{x,c,i}\|^2+(1-\nu\mu_y)
\Big[\mu_x \notag
\\
&+ \frac{(L+L_f)\mu^2_x}{2}\Big] 
\mathbb{E}\|{g}_{x,c,i}\|^2  - \frac{\mu_y}{2}(1 - L_f\mu_y)
\mathbb{E}\|{g}_{y,c,i}\|^2
 \notag\\ &+{\mu_y}L_f^2\mathbb{E}\|\boldsymbol{x}_{c, i+1}-\boldsymbol{x}_{c, i}\|^2
+ \mu_y\mathbb{E}\| \nabla_{y}J(\boldsymbol{x}_{c, i}, \boldsymbol{y}_{c, i}) - {g}_{y,c,i}
\|^2 \notag \\ &  +
5{(1- \nu \mu_y)(L + L_f)\mu^2_x\sigma^2} \underset{k}{\max} \  p^2_k + 5L_f\mu^2_y \sigma^2 \underset{k}{\max} \ p^2_k    \notag
\end{align}
We further choose $\frac{2\mu_xL_f^2}{\nu} \le 
\frac{\nu\mu_y}{2},\mu_y \le \frac{1}{\nu},
\mu^2_x \le \frac{L_f\mu^2_y}{L+L_f}$.
We have $(1 - \nu\mu_y)(1 + \frac{2\mu_x L_f^2}{\nu}) \le 1- \frac{\nu\mu_y}{2}$,
and inequality \eqref{eq:recursionforbn3} 
becomes 
\begin{align}\label{eq:recursionforbn4}
\mathbb{E}&[{d}_{c}(i+1)] \\
 \le&  (1 - \frac{\nu\mu_y}{2})\mathbb{E}[{d}_{c}(i)]
+
\frac{3\mu_x(1-\nu\mu_y)}{2}\mathbb{E}\|\nabla_{x}J(\boldsymbol{x}_{c, i}, \boldsymbol{y}_{c, i}) \notag\\&-
{g}_{x,c,i}\|^2+(1-\nu\mu_y)\Big[\mu_x + \frac{(L+L_f)\mu^2_x}{2}\Big]
\mathbb{E}\|{g}_{x,c,i}\|^2 \notag \\
& - \frac{\mu_y}{2}(1 - L_f\mu_y)
\|{g}_{y,c,i}\|^2
+ {\mu_y}L_f^2\mathbb{E}\|\boldsymbol{x}_{c, i+1}-\boldsymbol{x}_{c, i}\|^2
\notag\\&+ \mu_y\mathbb{E}\| \nabla_{y}J(\boldsymbol{x}_{c, i}, \boldsymbol{y}_{c, i}) - {g}_{y,c,i}
\|^2     + 10L_f\mu^2_y \sigma^2 \underset{k}{\max} \ p^2_k \notag
\end{align}
Choosing $\mu_y < \frac{1}{L_f}$ and
iterating \eqref{eq:recursionforbn4} from $i$ to $0$, we get 
\begin{align} \label{eq:recursionforbn5}
\mathbb{E}&[{d}_{c}(i)] \notag
\\
\le & (1 - \frac{\nu\mu_y}{2})^{i}{d}_{c}(0)
+\sum_{j=0}^{i-1} 
(1 - \frac{\nu\mu_y}{2})^{i-1-j}\Bigg\{
\frac{3\mu_x(1-\nu\mu_y)}{2}
\notag\\&\times \mathbb{E}\|\nabla_{x} J(\boldsymbol{x}_{c, j}, \boldsymbol{y}_{c, j})-
{g}_{x,c,j}\|^2+(1-\nu\mu_y)\Big[\mu_x  
 \notag \\&+ \frac{(L+L_f)\mu^2_x}{2}\Big]\mathbb{E}\|{g}_{x,c,j}\|^2 +
{\mu_y}L_f^2 \mathbb{E}\|\boldsymbol{x}_{c, j+1}-\boldsymbol{x}_{c, j}\|^2  \notag \\& + \mu_y\mathbb{E}\| \nabla_{y}J(\boldsymbol{x}_{c, j}, \boldsymbol{y}_{c, j}) - {g}_{y,c,j}
\|^2 
+ 10L_f\mu^2_y \sigma^2 \underset{k}{\max} \ p^2_k\Bigg\}
\notag\\&
-\frac{\mu_y}{2}(1 - L_f\mu_y)
\sum_{j=0}^{i-1} 
(1 - \frac{\nu\mu_y}{2})^{i-j-1}
\mathbb{E}\|{g}_{y,c,j}\|^2 \notag
\end{align}
\begin{align}
\overset{(a)}{\le} & (1 - \frac{\nu\mu_y}{2})^{i}{d}_{c}(0)
+\sum_{j=0}^{i-1} 
(1 - \frac{\nu\mu_y}{2})^{i-1-j}\Bigg\{
\frac{3\mu_x(1-\nu\mu_y)}{2}
\notag\\&\times \mathbb{E}\|\nabla_{x} J(\boldsymbol{x}_{c, j}, \boldsymbol{y}_{c, j})-
{g}_{x,c,j}\|^2+(1-\nu\mu_y)\Big[\mu_x  \notag
\\&+ \frac{(L+L_f)\mu^2_x}{2}\Big]\mathbb{E}\|{g}_{x,c,j}\|^2 +
{\mu_y}L_f^2 \mathbb{E}\|\boldsymbol{x}_{c, j+1}-\boldsymbol{x}_{c, j}\|^2 \notag \\& + \mu_y\mathbb{E}\| \nabla_{y}J(\boldsymbol{x}_{c, j}, \boldsymbol{y}_{c, j}) - {g}_{y,c,j}
\|^2\Bigg\} + \frac{20L_f\mu_y\sigma^2}{\nu } \underset{k}{\max} \  p^2_k
\notag\\&
-\frac{\mu_y}{2}(1 - L_f\mu_y)
\sum_{j=0}^{i-1} 
(1 - \frac{\nu\mu_y}{2})^{i-j}
\mathbb{E}\|{g}_{y,c,j}\|^2
\end{align}
where  $(a)$ is due to
$
(1 - \frac{\nu\mu_y}{2})^{i-j-1}
\|{g}_{y,c,j}\|^2
\ge 
(1 - \frac{\nu\mu_y}{2})^{i-j}
\|{g}_{y,c,j}\|^2
$ for any $j \le i-1$.
Averaging \eqref{eq:recursionforbn5}
over iterations, we get 
\begin{align}\label{bnPL}
\frac{1}{T} &\sum_{i=0}^{T-1} \mathbb{E}[ {d}_{c}(i)] \notag
  \\
   \overset{(a)}{\le} &
  \frac{2}{\nu\mu_y T}{d}_{c}(0)
  + 
\frac{1}{T}
\sum_{i=0}^{T-1}\Bigg\{
  \frac{3\mu_x(1-\nu\mu_y)}{\nu\mu_y}
  \mathbb{E}\|\nabla_{x} J(\boldsymbol{x}_{c, i}, \boldsymbol{y}_{c, i}) \notag\\&-
{g}_{x,c,i}\|^2 + 
(1-\nu\mu_y)\Big[\frac{2\mu_x}{\nu\mu_y} + \frac{(L+L_f)\mu^2_x}{\nu\mu_y}\Big]\mathbb{E}\|{g}_{x,c,i}\|^2 +\notag\\
& \frac{2L_f^2}{\nu }\mathbb{E}\|\boldsymbol{x}_{c, i+1}-\boldsymbol{x}_{c, i}\|^2
+\frac{2}{\nu} \mathbb{E}\| \nabla_{y}J(\boldsymbol{x}_{c, i}, \boldsymbol{y}_{c, i}) - {g}_{y,c,i}\|^2 \Bigg\}
\notag\\
&
- \frac{\mu_y}{2T}(1 - L_f\mu_y)
\sum_{i=1}^{T-1} \sum_{j=0}^{i-1}
(1 - \frac{\nu\mu_y}{2})^{i-j}
\mathbb{E}\|{g}_{y,c,j}\|^2 \notag \\
& +\frac{20L_f\mu_y\sigma^2}{\nu } \underset{k}{\max} \  p^2_k 
\end{align}
where (a) follows from the following inequality
\begin{align}\label{basicinequality}
\sum_{i= 1}^{T-1} \sum_{j = 0}^{i-1}
    q^{i-1-j}a_j
    = 
    \sum_{i=0}^{T-2}a_i  
    (\sum_{j=0}^{T-2-i} q^j)\le& 
    \frac{1}{1-q}\sum_{i=0}^{T-2}a_i \notag \\
    \le& \frac{1}{1-q}\sum_{i=0}^{T-1}a_i
\end{align}
here, $0<q <1$ is a constant, $\{a_i\}$
is a positive sequence.
\bibliographystyle{IEEEbib}
{\footnotesize
\bibliography{refs}
}
\end{document}